\documentclass[12pt]{article}

\usepackage[inline,shortlabels]{enumitem}
\usepackage{float}
\usepackage{mathtools}
\usepackage{subdepth}
\usepackage{comment}
\usepackage{graphicx}
\usepackage[round]{natbib}
\usepackage{algpseudocode}
\usepackage{tablefootnote}
\usepackage[margin=1in]{geometry}
\usepackage{tikz}
\usetikzlibrary{arrows.meta}
\usetikzlibrary{decorations.markings}
\usepackage[linesnumbered]{algorithm2e}
\RestyleAlgo{ruled}
\usetikzlibrary{plotmarks}
\usetikzlibrary{shapes,decorations,arrows,calc,arrows.meta,fit,positioning}
\usetikzlibrary{shapes.geometric}
\tikzset{
font={\fontsize{7pt}{12}\selectfont},
every node/.style={scale=1.3},
square/.style={regular polygon,regular polygon sides=4},
    -Latex,auto,node distance =1 cm and 1 cm,semithick,
    state/.style ={ellipse, draw, minimum width = 0.7 cm},
    block/.style = {square, draw, inner sep=0cm,minimum size=8mm},
    free/.style = {circle, draw, inner sep=0cm,minimum size=6mm},
    bidirected/.style={Latex-Latex,dashed},
    el/.style = {inner sep=2pt, align=left, sloped}
}\usepackage[english]{babel}
\usepackage{longtable}
\usepackage{nccmath}
\usepackage{color}
\usepackage{mathtools}
\usepackage{amssymb,amsmath,amsthm}
\usepackage{multirow,multicol,makecell,booktabs}
\usepackage[titletoc,title]{appendix}
\usepackage{authblk}
\usepackage{bbm}
\usepackage{setspace}
\usepackage{dsfont}
\usepackage[OT1]{fontenc}
\usepackage{subcaption}
\usepackage{refcount}
\usepackage{booktabs}
\graphicspath{ {../simulations/01_init_manuscript/graphs/} }
\usepackage{xr}
\usepackage[colorlinks,citecolor=blue,urlcolor=blue]{hyperref}

\newtheorem{remark}{Remark}
\newtheorem{proposition}{Proposition}
\newtheorem{theorem}{Theorem}
\AtEndDocument{\refstepcounter{theorem}\label{finalthm}}
\AtEndDocument{\refstepcounter{equation}\label{finaleq}}

{
  \theoremstyle{definition}
  \newtheorem{assumption}{}
}
{
  \theoremstyle{definition}
  
}
{
  \theoremstyle{definition}
  
}

\newtheorem{lemma}{Lemma}
\AtEndDocument{\refstepcounter{lemma}\label{finallemma}}
\newtheorem{coro}{Corollary}
\newtheorem{definition}{Definition}

 \DeclareMathOperator{\var}{\mathsf{Var}}

\renewcommand{\P}{\mathsf{P}}

\newcommand{\Q}{\mathsf{Q}}
\newcommand{\F}{\mathsf{F}}

\let\sec\S 

\renewcommand{\S}{\mathsf{S}}

\newcommand{\indep}{\mbox{$\perp\!\!\!\perp$}} 
 \newcommand{\dd}{\,\mathrm{d}}

\newcommand{\Zp}{Z^\pi}

\newcommand{\one}{\mathds{1}}
 
\newcommand{\E}{\mathsf{E}}
\renewcommand{\P}{\mathsf{P}}

\usepackage{accents}

\DeclarePairedDelimiterX{\norm}[1]{\lVert}{\rVert}{#1}

\pgfdeclarelayer{background}
\pgfsetlayers{background,main}
\usetikzlibrary{arrows,positioning}
\tikzset{
>=stealth',
punkt/.style={
rectangle,
rounded corners,
draw=black, very thick,
text width=6.5em,
minimum height=2em,
text centered},
pil/.style={
->,
thick,
shorten <=2pt,
shorten >=2pt,}
}
\newcommand{\Vertex}[3]% pos, name
{\node[minimum width=0.6cm,inner sep=0.05cm] (#2) at (#1) {#3};
}
\newcommand{\Vertexr}[3]% pos, name
{\node[rectangle, draw, minimum width=0.6cm,inner sep=0.05cm] (#2) at (#1) {#2};
}
\newcommand{\ArrowR}[3]%
{ \begin{pgfonlayer}{background}
\draw[->,#3] (#1) to[bend right=30] (#2);
\end{pgfonlayer}
}
\newcommand{\ArrowLW}[3]%
{ \begin{pgfonlayer}{background}
\draw[->,#3] (#1) to[bend left=30] (#2);
\end{pgfonlayer}
}

\newcommand{\ArrowL}[3]%
{ \begin{pgfonlayer}{background}
    \draw[->,#3] (#1) to[bend left=45] (#2);
  \end{pgfonlayer}
}

\newcommand{\EdgeL}[3]%
{ \begin{pgfonlayer}{background}
\draw[dashed,#3] (#1) to[bend right=-45] (#2);
\end{pgfonlayer}
}

\newcommand{\Arrow}[3]%
{ \begin{pgfonlayer}{background}
\draw[->,#3] (#1) -- +(#2);
\end{pgfonlayer}
}

\allowdisplaybreaks
\pgfarrowsdeclare{arcs}{arcs}{...}
{
  \pgfsetdash{}{0pt} % do not dash
  \pgfsetroundjoin   % fix join
  \pgfsetroundcap    % fix cap
  \pgfpathmoveto{\pgfpoint{-5pt}{5pt}}
  \pgfpatharc{180}{270}{5pt}
  \pgfpatharc{90}{180}{5pt}
  \pgfusepathqstroke
}
\newcommand{\ArrowB}[3]%
{ \begin{pgfonlayer}{background}
    \draw[|-arcs,line width=0.4mm,shorten <= 0.3cm,shorten >= 0.3cm,#3] (#1) -- +(#2);
  \end{pgfonlayer}
}

\newcommand{\titlepaper}{General targeted machine learning for modern causal mediation analysis}

\date{\today}

\author[1,*]{Richard Liu}
\author[2]{Nicholas T. Williams}
\author[2]{Kara E. Rudolph}
\author[1]{Iv\'an D\'iaz}

\affil[1]{\small Division of Biostatistics, Department of Population
  Health, New York University Grossman School of Medicine, USA}

\affil[2]{\small Department of Epidemiology, Mailman School of Public Health, Columbia University, New York, NY, USA.}

\affil[*]{\small Corresponding Author. Email: \href{mailto:richard.l@nyu.edu}{richard.l@nyu.edu}}
\title{\titlepaper}

\begin{document}
\maketitle
% \doublyspacing
% \linespread{2}

\begin{abstract}
 
Causal mediation analyses investigate the mechanisms through which causes exert their effects, and are therefore central to scientific progress. The literature on the non-parametric definition and identification of mediational effects in rigorous causal models has grown significantly in recent years, and there has been important progress to address challenges in the interpretation and identification of such effects. However, statistical methodology for non-parametric estimation has lagged, with few or no methods available for tackling non-parametric estimation in the presence of multiple, continuous, or high-dimensional mediators. In this paper we propose an all-purpose one-step estimation algorithm that can be coupled with machine learning in mediation studies that use any of six common mediation parameters (natural direct and indirect effects, randomized interventional effects, separable effects, organic direct and indirect effects,  recanting twin effects, and decision theoretic effects). The estimators build on methods for double machine learning, including a re-parameterization of the identification formulas in terms of sequential regressions, a first order non-parametric von-Mises approximation of the first-order bias of a plug-in estimator, and Riesz learning for estimation of nuisance parameters. We show that the proposed one-step estimators have desirable properties, such as $\sqrt{n}$-convergence and asymptotic normality. 
We illustrate the properties of our methods in a simulation study and demonstrate its use on real data to estimate the extent to which pain management practices mediate the total effect of having a chronic pain disorder on opioid use disorder. We provide an R package (on CRAN) implementing our methods publicly available at \url{https://github.com/nt-williams/crumble}.

\end{abstract}
% \keywords{}    \node [label={[xshift=1.0cm, yshift=0.3cm]Label}] {Node};
% \newpage

\newpage 

\section{Introduction}\label{sec:intro}

\subsection{Prior literature}
% \textit{Paragraph 1: Discussion of mediation analysis and its importance in science.}
Causal mediation analyses seek to investigate the extent through which the effect of an exposure on an outcome operates through effects of the exposure on intermediate variables. Recent decades have seen increased attention to the development of methodology for the definition, identification, and estimation of effects for mediation analysis, as well as their increased application in various scientific fields. Particularly, the definition and identification of effects that are guaranteed to measure the \textit{mechanisms} through which the effect operates has been the subject of considerable debate and methodological work \citep[e.g.,][]{robins2003semantics,pearl2010introduction,robins2010alternative,miles2023causal,diaz2024non}. 

Natural direct and indirect effects \citep[NDE and NIE;][]{RobinsGreenland92,pearl2001direct} are one of the most widely known definitions of causal mediation parameters with an agreed upon mechanistic interpretation. Broadly, the NDE measures the effect that operates independent of a mediator through counterfactual variables in which the effect through the mediator is disabled, whereas the NIE measures effects through the mediator by considering counterfactuals that disable all effects that operate independently of it. In spite of their scientific importance, desirable interpretation, and widespread use, the NDE and NIE are seldom identified from observed data, because their identification relies on so-called \textit{cross-world} counterfactual assumptions---independence assumptions between counterfactual variables indexed by distinct interventions on the exposure. Philosophically, cross-world assumptions are problematic because they cannot be tested empirically nor enforced by design, which means that conclusions from an analysis using the NIE and NDE are not falsifiable. 
Practically, the cross-world assumption is guaranteed not to hold in the presence of variables caused by the treatment that are common causes of both the mediator and the outcome \citep{avin2005identifiability}, regardless of whether such variables are measured or not. These variables, often referred to as \textit{intermediate confounders}, are pervasive in scientific applications. Although \cite{vo2024recanting} proposed a test to confirm whether a given variable $Z$ is an intermediate confounder, this test cannot rule out the existence of unmeasured intermediate confounders that would invalidate the cross-world assumption. 

In an attempt to address the above limitations of the NDE and NIE, several alternative causal parameters have been proposed. In this article we discuss five such definitions, but we note that our review is not exhaustive. First, randomized interventional direct and indirect effects \citep[RIDE and RIIE;][]{van2008direct,vanderweele2014effect,vansteelandt2017interventional,vanderweele2017mediation,diaz2023efficient} define direct and indirect effects through interventions that set the mediator to a random draw based on the distributions of the counterfactual mediators under treatment and control. Randomized interventional effects are identifiable in the presence of intermediate confounders under the assumption that all confounders of the exposure-outcome, exposure-mediator, and mediator-outcome relations are measured. However, it was recently shown that randomized interventional effects may not necessarily measure mechanistic relations, as the RIIE could be nonzero even in a situation where some population units experience an exposure-mediator effect, other units experience a mediator-outcome effect, but no unit experiences both \citep{miles2023causal}.  
Second, in a series of articles, \cite{didelez2006direct,geneletti2007identifying} and \cite{dawid2021decision} proposed a decision-theoretic approach to mediation analysis (and more generally to causal inference) that uses probabilistic definitions for the relevant interventions, avoiding reliance on counterfactual arguments, and therefore avoiding reliance on cross-world independence assumptions. Third, a recent strand of the causal inference literature has addressed the definition and identification of path-specific effects and effects under competing events using so-called separable effects \citep{robins2010alternative,didelez2019defining,stensrud2021generalized,robins2022interventionist,
stensrud2022separable}. These effects avoid cross-world definitions by requiring researchers to posit the existence of separable components of the exposure that exert independent effects on the outcome and the mediator. Separable effects are then defined through distinct interventions on those components, thus avoiding cross-world statements. \cite{vo2024recanting} generalized separable effects to allow for the definition and identification of mediational effects in the presence of intermediate confounders. Fourth, \cite{diaz2024non} and \cite{vo2024recanting} recently introduced a four-way decomposition of the average treatment effect into each of the path-specific effects involved, using randomized versions of the recanting witnesses. Unlike some of the previous approaches, these path-specific effects satisfy a property known as the \textit{path-specific sharp null criterion} \citep{miles2023causal}, which guarantees that the effects are zero whenever no individual in the population experiences an effect through the corresponding path. Lastly, a series of articles discusses organic direct and indirect effects \citep{lok2015organic,lok2016defining,lok2019causal,lok2021causal}, which rely on the characterization of an external intervention to the causal system that would yield specific modifications to the mediator distribution.

\subsection{Our contribution}

In spite of important progress in the non-parametric definition and identification of mediational parameters, \textit{estimators} for many approaches remain underdeveloped, and there are few non-parametric solutions\footnote{See \cite{van2008direct, zheng2012targeted,xia2023identification} for estimators of the NDE/NIE. See \cite{diaz2021nonparametric,benkeser2021nonparametric,rudolph2024practical} and \cite{zheng2017longitudinal,diaz2023efficient, wang2023targeted} for randomized interventional effects for cross-sectional and longitudinal settings, respectively.} with only a handful that can handle continuous or high-dimensional mediators in specific settings.\footnote{To our knowledge, only the approach of \cite{rudolph2024practical} for randomized interventional effects allows for high-dimensional mediators and intermediate confounders.} Methods for mediation analyses that can handle continuous or high-dimensional mediators are paramount in contemporary research fields such as the study of so-called omics, (e.g., the microbiome is often considered a mediator of effects on health \citep{sohn2019compositional,wang2020estimating}), studies of fairness in machine learning \citep{nabi2024fair}, and neuroscience \citep{chen2018high}, among many others. An additional limitation of all of the above approaches for the definition, identification, and estimation of direct and indirect effects in mediation analyses is that the literature has almost exclusively focused on binary exposures, although some approaches for continuous and/or multivariate exposures have been developed based on stochastic interventions and modified treatment policies \citep{diaz2020causal,hejazi2023nonparametric,gilbert2024identification}. In contrast, we resolved the estimation issue by making two main contributions. First, we developed non-parametric double machine learning estimators for all six mediation parameters listed in the abstract for the case of binary treatments in the presence of continuous or multivariate, mediators and intermediate confounders. Second, we proposed two approaches (based on randomized interventional effects, recanting-twins effects) for defining mediation parameters in the presence of continuous or multivariate treatments, and proposed estimation methods that can handle continuous or multivariate, mediators and intermediate confounders.

Our estimators are based on recently popularized frameworks that estimate and/or correct for the first-order bias of a plug-in machine learning estimator \citep{vanderLaanRose11,vanderLaanRose18,chernozhukov2018double}. Central to these frameworks is the concept of a \textit{canonical gradient} or \textit{efficient influence function} (EIF), which characterizes the first-order bias of plug-in estimators as well as the non-parametric efficiency bound \citep{mises1947asymptotic, pfanzagl1982contributions, Bickel97}. A key challenge in mediation analysis with continuous or high-dimensional mediators is that estimation of the canonical gradient requires estimation of densities of the mediator as well as integrals with respect to such densities. We address challenges in the estimation of multivariate or high-dimensional densities on the mediator through the use of so-called Riesz learning \citep{chernuzhukov2022riesznet}, and address integration with respect to those densities through parameterizing the integrals as conditional expectations using stratified random permutations of the observed variables as in \cite{doran2014permutation}.  Definitions of mediation parameters for continuous or multivariate exposures leverage \textit{modified treatment policies} \citep{Haneuse2013, sani2020identification, diaz2021nonparametricmtp, diaz2022causalcomp}, also called \textit{dynamic interventions that depend on the natural value of treatment} \citep{robins2004effects, richardson2013single, young2014identification}.

The rest of the paper is organized as follows. In \sec\ref{sec:general}, we introduce general identification formulas for mediation analysis under a binary exposure, where we use the fact that the identification formulas for the six causal parameters for mediation analysis with binary exposures listed in the abstract can be recovered from two fundamental statistical functionals. In \sec\ref{sec:theory}, we introduce the semi-parametric efficiency theory these two functionals. In \sec\ref{sec:estima}, we use the developed theory to introduce estimation strategies. We prove desirable properties of the estimators such as asymptotic linearity, weak convergence, and double robustness. In \sec\ref{sec:mtp} we present definitions for non-binary treatments using modified treatment policies and introduce the corresponding estimators. In \sec\ref{sec:simula}, we present numerical illustrations of the performance of the proposed approach in simulated datasets. An illustrative real data example investigating the effects of chronic pain disorder on opioid use disorder is given in \sec\ref{sec:ilustra}. A discussion about this manuscript and future work is given in \sec\ref{sec:discussion}.

% \textit{Paragraph 1: Discussion of mediation analysis and its importance in science.}

\section{General identification formulas for mediation analysis with binary exposures} \label{sec:general}
\begin{figure}[!htb]
  \centering
  \begin{tikzpicture}
    \tikzset{line width=1pt, outer sep=0pt,
      ell/.style={draw,fill=white, inner sep=2pt,
        line width=1pt}};
    \node[circle, draw, name=a, ]{$A$};
    \node[circle, draw, name=z, right = 10mm of a]{$Z$};
    \node[circle, draw, name=m, right = 10mm of z]{$M$};
    \node[circle, draw, name=y, right = 10mm of m]{$Y$};
    \node[circle, draw, name=w, above right = 7mm and 3mm of z]{$W$};

    \draw[->](a) to (z);
    \draw[->](w) to (z);
    \draw[->](z) to (m);
    \draw[->](a) to[out=-45,in=225] (y);
    \draw[->](a) to[out=-45,in=225] (m);
    \draw[->](z) to[out=-45,in=225] (y);
    \draw[->](w) to (y);
    \draw[->](m) to (y);
    \draw[->](w) to (a);
    \draw[->](w) to (m);
  \end{tikzpicture}
  \caption{Causal DAG}
  \label{fig:dag}  
\end{figure}
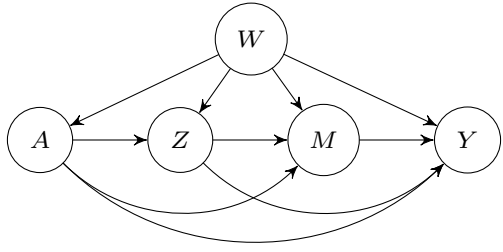

In this section we show that the identification formulas for the six approaches to mediation discussed in the abstract and introduction can be recovered from just two generalized statistical estimands. This fact will then be used to propose a multi-purpose estimator for generalized mediation analyses. We will first introduce some notation. Let $A \in \{0, 1\}$ denote a binary treatment or exposure variable (we discuss effects for non-binary exposures in \sec\ref{sec:mtp}), let $Y$ denote a continuous or binary outcome, let $M$ denote a vector of mediators, let $Z$ denote a vector of intermediate confounders, and let $W$ denote a vector of observed covariates. Let $X = (W, A, Z, M, Y)$ represent a random variable with distribution $\P$. Let $\P_n$ denote the empirical distribution of a sample of independent and identically distributed (i.i.d.) observations $X_1, \ldots, X_n$, and let $\P f = \int f(x)\dd \P (x) $ for a given function $f(x)$. Let $\E_\P$ be the expectation with respect to $\P$. When there is no ambiguity we will denote $\E = \E_\P$. The causal models used in the original articles that define the causal mediational parameters we consider are varied (e.g., not all use counterfactuals), but they all have the same representation in terms of the directed acyclic graph (DAG) given in Figure~\ref{fig:dag}, where sometimes the variable $Z$ is present and sometimes it is not.  
Our methods will rely on the availability of a random variable $\Zp$ satisfying 
\begin{equation}\label{eq:defZp}
    \Zp \mid (A, W) \sim Z \mid (A, W) \text{ and } 
    \Zp \indep M \mid (A, W),
\end{equation}
where $A \sim B$ means equality in distribution. In this section as well as in \sec\ref{sec:estima}, we assume that an i.i.d. sample $(\Zp_1,\ldots,\Zp_n)$ of such a random variable is given together with $X_1, \ldots, X_n$, so that the observed data for unit $i$ can be augmented as $(W_i, A_i, Z_i, \Zp_i, M_i, Y_i)$. In practice, $\Zp_i$ will be constructed through a permutation of $Z_i$ which we discuss in \sec\ref{sec:genzp}.%; in \sec\ref{} we will show that this construction does not affect the asymptotic properties of our estimators. 

Let $\theta(A,Z,M,W)=\E[Y\mid A,Z,M,W]$. For values $(a_1,\ldots, a_4) \in \{0,1\}^4$, we define
{\footnotesize
\begin{equation}
\begin{aligned}
    \psi^R(a_1,a_2,a_3,a_4)  & = \E\Bigg[\E\bigg[\E\Big[\E \big[ \theta (a_1, \Zp, M, W ) \,\big|\, A=a_2,M, W\big ]\,\Big|\, A=a_3, Z, W\Big]\,\bigg|\, A=a_4, W\bigg]\Bigg],\\
    \psi^N(a_1,a_2,a_3)   & = \E \bigg[\E \Big[ \E \big[ \theta(a_1, Z, M, W) \,\big|\, A = a_2, Z, W\big] \,\Big|\, A = a_3, W\Big]\bigg],
\end{aligned}\label{eq:defpsi}    
\end{equation}
}%
where the superscript $N$ stands for natural and the superscript $R$ stands for randomized, accounting for the fact that we introduce the permutation $\Zp$. In Proposition \ref{prop:equivalence} below we show that the identifying functionals for the six mediation frameworks described in the abstract can be recovered from just two parameters $\psi^R$ and $\psi^N$ defined above. The introduction of the permuted variable $\Zp$ allows us to write integrals of the type $\int f(a_1, z, m, w)\dd\P(z\mid a_2, w)$, which may be hard to estimate numerically if $Z$ is continuous or high-dimensional, as conditional expectations  $\E[f(a_1,\Zp,M,W)\mid A=a_2, M=m, W=w]$, which can be estimated using flexible regression techniques from the statistical and machine learning literature. 

In Table~\ref{tab:params} below we introduce abbreviations for each of the corresponding causal effects, and provide one reference for each where the corresponding definitions in terms of causal models and proofs for identification can be found. Of the approaches listed in Table~\ref{tab:params}, only the first two do not allow for intermediate confounders (i.e., these two approaches assume there is no variable $Z$ in the DAG in Figure~\ref{fig:dag}), and only the last two allow for a decomposition of the average treatment effect into path-specific effects. The parameters referred to as ``indirect effects'' measure all effects of $A$ on $Y$ that operate through $M$ (i.e., $A\to M\to Y$ and $A\to Z\to M\to Y$), and the parameters referred to as ``direct effects'' measure all effects of $A$ on $Y$ that operate independently of $M$ (i.e., $A\to Z\to Y$ and $A\to Y$). For approaches that allow a decomposition into path-specific effects, we introduce the notation $P_j$ to refer to each path, where $P_1: A\rightarrow Y$,
$P_2: A \rightarrow Z \rightarrow Y$, $P_3: A \rightarrow Z \rightarrow M \rightarrow Y$, and
$P_4: A \rightarrow M \rightarrow Y$. A comprehensive discussion of the causal models, definitions of mediational effects, and identification assumptions, is out of the scope of this paper and is given in the references provided in \sec\ref{sec:intro} and in Table~\ref{tab:params}, we refer the readers to those references for discussion.
\begin{table}[H]
\centering{\footnotesize
\begin{tabular}{c|l|l}\hline
           Abbreviation  &  Parameter & Reference  \\\hline
           NDE/NIE  & Natural direct and indirect effects & \cite{pearl2001direct} \\
           DTDE/DTIE & Decision theoretic direct and indirect effects & \cite{geneletti2007identifying}\\
           ODE/OIE & Organic direct and indirect effects & \cite{lok2015organic} \\
           RIDE/RIIE & Randomized interventional direct and indirect effects & \cite{vanderweele2014effect}\\
           RT$_j:j=1,2,3,4$ & Path-specific effects based on recanting twins for path $P_j$& \cite{diaz2024non}\\
            SE$_j:j=1,2,3,4$ & Path-specific effects based on separable effects for path $P_j$&\cite{vo2024recanting}\\\hline
        \end{tabular}}%
        \caption{Abbreviations and references with the definitions of the causal effects and their identification theorems for the six mediation frameworks considered in this work}
        \label{tab:params}
    \end{table}
\begin{proposition}[Identification of parameters for six approaches to causal mediation]\label{prop:equivalence}
The identification formulas for the causal parameters listed in Table~\ref{tab:params} and provided in the corresponding references can be written in terms of $\psi^N$ and $\psi^R$ as:
{\footnotesize
\begin{align*}
    \operatorname{NDE} &= \psi^N(1,0,0) - \psi^N(0,0,0), &\operatorname{NIE} &= \psi^N(1,1,1) - \psi^N(1,0,0),\\ 
        \operatorname{DTDE} &= \psi^N(1,0,0) - \psi^N(0,0,0), &\operatorname{DTIE} &= \psi^N(1,1,1) - \psi^N(1,0,0),\\  
        \operatorname{ODE} &= \psi^N(1,0,1) - \psi^N(0,0,0), &\operatorname{OIE} &= \psi^N(1,1,1) - \psi^N(1,0,1),\\  
        \operatorname{RIDE} &= \psi^R(1,1,0,0) - \psi^R(0,0,0,0), &\operatorname{RIIE} &= \psi^R(1,1,1,1) - \psi^R(1,1,0,0),
\end{align*}}%, 
as well as
{\footnotesize
\begin{align*}
    \operatorname{RT}_1 &= \psi^N(1,1,1)-\psi^N(0,1,1), &\operatorname{RT}_2 &= \psi^R(0,1,1,1) - \psi^R(0,0,1,1),\\
    \operatorname{RT}_3 &= \psi^R(0,0,1,1) - \psi^R(0,0,1,0), &\operatorname{RT}_4 &= \psi^N(0,1,0) - \psi^N(0,0,0),
\end{align*}}%
and
{\footnotesize
\begin{align*}
    \operatorname{SE}_1 &= \psi^N(1,1,1)-\psi^N(0,1,1), &\operatorname{SE}_2 &= \psi^R(0,1,1,1) - \psi^R(0,0,1,1),\\
    \operatorname{SE}_3 &= \psi^R(0,0,1,1) - \psi^R(0,0,1,0), &\operatorname{SE}_4 &= \psi^N(0,1,0) - \psi^N(0,0,0).
\end{align*}}%
\end{proposition}
\begin{remark}
The recanting twins effects discussed in \cite{vo2024recanting} are generally different from the natural path-specific effects. However, recanting twin effects have desirable properties that recover features of natural path-specific effects, such as 1) monotonicity, meaning they increase or decrease in alignment with the natural path-specific effects, 2) have the same sign with the natural path-specific effects, and 3) recover natural path-specific effects in the absence of intermediate confounding.
\end{remark}

A proof of Proposition \ref{prop:equivalence} is shown in the supplementary materials. Having established the importance and generality of the parameters $\psi^N$ and $\psi^R$ for mediation analysis, we will proceed to develop general estimation methods for those parameters that can be used with high-dimensional $W$, $Z$, and $M$ variables. In what follows, we ease notation by omitting $(a_1,a_2,a_3,a_4)$ when referring to $\psi^N$ and $\psi^R$, but the dependence of the parameters on these values will remain implicit.  In order to develop non-parametric estimators that can leverage machine learning to fit the regressions involved in the definitions of $\psi^N$ and $\psi^R$, we will use techniques from the literature in semi-parametric efficiency theory, which require the study of certain properties of the parameters  $\psi^N$ and $\psi^R$ seen as functionals $\psi^N(\P)$ and $\psi^R(\P)$ that map a probability distribution $\P$ in the non-parametric model $\cal P$ into a real number given by the formulas in (\ref{eq:defpsi}). The following section provides a study of the required elements. 

\section{Semi-parametric efficiency theory} \label{sec:theory}

The estimators we propose will be based on one-step first-order bias corrections to plug-in machine learning estimators based on the formulas in (\ref{eq:defpsi}). To motivate the developments that follow, consider estimators of the formulas in (\ref{eq:defpsi}) constructed, for example, by estimating the probability distribution $\P$ and plugging it into the formulas, or by estimating the sequential regressions involved using predictions from the prior regressions in the sequence as pseudo-outcomes. When such plug-in estimators are constructed based on estimators from parametric models, the Delta method allows one to prove that the distribution of the $\sqrt{n}$-scaled estimator may be approximated by a Gaussian random variable for large $n$, allowing the analytical computation of uncertainty measures such as confidence intervals. The Gaussian distribution, however, will be centered at the wrong value if the models are incorrectly specified, with the alarming consequence that the coverage of a confidence interval will contain the true parameter value with zero probability as sample size grows. To address this model misspecification bias, the formulas in  (\ref{eq:defpsi}) may also be estimated using data-adaptive methods from the machine learning literature, which may allow for additional flexibility in automatically finding interactions, non-linearities, etc. However, no theory exists that allows the analysis of the sampling distributions of general plug-in machine learning estimators. Instead, recent methods for machine learning in causal inference such as targeted learning \citep{vanderLaanRose11,vanderLaanRose18} or double machine learning \citep{chernozhukov2018double} rely on characterizing and correcting for the first-order bias of the plug-in machine learning estimators. The characterization of the bias is given by a so-called first-order von Mises expansion \citep{mises1947asymptotic,fernholz1983mises,robins2009quadratic}, defined below. In what follows we use $\psi(\P)$ to refer to a general statistical parameter viewed as a map from the non-parametric models to the real line.

\begin{definition}[First-order von Mises expansion]
    Let $\psi : \mathcal P \to \mathbb R$ be a functional from the non-parametric model to the real line. The functional admits a first-order von Mises expansion if it satisfies: 
    \begin{equation}
        \psi(\F) - \psi(\P) = -\E_{{\P}}[\varphi (X; \F)] + R_2(\F, \P) \label{eq:von-mises}
    \end{equation}
    for any $\F,\P\in\cal P$, where $\varphi(x; \F)$ is a function satisfying $\E_{\F}[\varphi (X; \psi(\F))]  = 0$ and $\E_{\F}[\varphi^2 (X; \psi(\F)) ]< \infty$, and the term $R_2(\F, \P)$ is  a second-order remainder term depending on products or squares of differences between $\F$ and $\P$. The function $\varphi$ is typically referred to as a gradient of the first-order expansion. 
\end{definition}
The above expansion provides the basis to analyze a plug-in machine learning estimator. For instance, assume that we have access to an estimate $\hat \P$ of the true data distribution $\P$, where  $\hat \P$ is obtained from data-adaptive methods. Then, Eq. (\ref{eq:von-mises}) provides the basis to analyze the error of the plug-in estimator $\psi(\hat\P)$ as
\[\psi(\hat\P) - \psi(\P) = -\E_{{\P}}[\varphi (X; \hat\P)] + R_2(\hat\P, \P).\]
Informally, if $\hat\P$ is a ``good enough'' estimator, the second-order term $R_2(\hat\P, \P)$ may be negligible, so that the error in estimation may be reduced to the first order term $-\E_{{\P}}[\varphi (X; \hat\P)]$. This error can then be estimated as the negative of the empirical average of $\varphi (X_i; \hat\P)$, which can then be added back to the plug in estimator $\psi(\hat\P)$. In the literature, this has been called a ``one-step'' estimator or a de-biased estimator, although the ``de-biased'' denomination is technically inaccurate as the above is an error analysis, not a bias analysis. (A bias analysis would consider the difference between the expectation of $\psi(\hat\P)$ and $\psi(\P)$, where the expectation is across multiple draws of $(X_1,\ldots,X_n)$.)

Theorem~\ref{theo:vonm} below provides a first-order von Mises expansion for $\psi^N$ and $\psi^R$ that can be used to construct a one-step estimator. To state the theorem, we introduce some additional notation. First, for $\psi^N$, consider a fixed value $(a_1,a_2,a_3)$ and define
{\small
\begin{equation} \label{nonprimetheta}
\begin{aligned}
\theta^N_3(a,z,m,w)&=\E(Y\mid A=a,Z=z,M=m,W=w), &     b^N_3(z,m,w;\theta_3^N)&=\theta^N_3(a_1,z,m,w),\\
    \theta^N_2(a,z,w)&=\E[b^N_3(Z,M,W;\theta_3^N)\mid A=a, Z=z, W=w],    & b^N_2(z,w;\theta_2^N)&=\theta^N_2(a_2,z,w),\\
        \theta^N_1(a,w) &= \E[b^N_2(Z,W;\theta_2^N)\mid A=a,W=w],     & b^N_1(w;\theta_1^N)&=\theta^N_1(a_3,w),\\
\end{aligned}
\end{equation}}%
as well as
{\small
\begin{equation}
\begin{aligned} \label{nonprimealpha}
         \alpha^N_1(a,w) & = \frac{\one(A=a_3)}{\P(a\mid w)},\\
        \alpha^N_2(a,z,w) & = \frac{\one(A=a_2)}{\P(a\mid w)}\frac{\P(z\mid A=a_3, w)}{\P(z\mid a, w)},\\
        \alpha^N_3(a,z,m,w) & = \frac{\one(A=a_1)}{\P(a\mid w)}\frac{\P(m\mid A=a_2, z, w)}{\P(m\mid a, z, w)}\frac{\P(z\mid A=a_3, w)}{\P(z\mid a, w)}.
\end{aligned}
\end{equation}}%
For $\psi^R$, consider a fixed value $(a_1,a_2,a_3,a_4)$ and define
{\small
\begin{equation} \label{primetheta}
\begin{aligned}
\theta^R_4(a,z,m,w)&=\E(Y\mid A=a,Z=z,M=m,W=w), &b^R_4(z,m,w;\theta_4^R)&=\theta^R_4(a_1,z,m,w),\\
\theta^R_{3}(a,m,w)&=\E[b^R_4(\Zp,M,W;\theta_4^R)\mid A=a, M=m, W=w], &b^R_{3}(m,w;\theta_{3}^R)&=\theta^R_{3}(a_2,m,w), \\
\theta^R_{2}(a,z,w)&=\E[b^R_{3}(M,W;\theta_{3}^R)\mid A=a, Z=z, W=w], & b^R_{2}(z,w;\theta_{2}^R)&=\theta^R_{2}(a_3,z,w), \\
\theta^R_{1}(a,w) &= \E[b^R_{2}(Z,W;\theta_{2}^R)\mid A=a,W=w], & b^R_{1}(w;\theta_{1}^R)&=\theta^R_{1}(a_4,w),  \\
\end{aligned}
\end{equation}}%
as well as
{\small
\begin{equation} \label{primealpha}
\begin{aligned}
\alpha^R_{1}(a,w) & = \frac{\one(A=a_4)}{\P(a\mid w)};\\
\alpha^R_{2}(a,z,w) & = \frac{\one(A=a_3)}{\P(a\mid w)}\frac{\P(z\mid A=a_4, w)}{\P(z\mid a, w)};\\
\alpha^R_{3}(a,m,w) & = \frac{\one(A=a_2)}{\P(a\mid w)}\frac{\int_z \P(m\mid A=a_3, z, w) \dd \P(z\mid A=a_4,w) }{\P(m\mid a, w)};\\
\alpha^R_4(a,z,m,w) & = \frac{\one(A=a_1)}{\P(a\mid w)}\frac{\P(z\mid A=a_2, w)}{\P(z\mid a, w)}\frac{\int_z \P(m\mid A=a_3, z, w) \dd \P(z\mid A=a_4,w) }{\P(m\mid a, z, w)}.
\end{aligned}
\end{equation}}%
Then we have the following result. 
\begin{theorem}[First-order von Mises expansions and efficiency bounds for $\psi^N$ and $\psi^R$] \label{theo:vonm} Let $\eta^N=(\theta_3^N, \theta_2^N, \theta_1^N, \alpha_3^N, \alpha_2^N, \alpha_1^N)$ and  $\eta^R=(\theta_3^R, \theta_3^R, \theta_2^R, \theta_1^R, \alpha_4^R, \alpha_3^R, \alpha_2^R, \alpha_1^R)$. The functions
{\small
\begin{align*}
    \varphi^{N}(X; \eta^N) & = \bar\varphi^{N}(X; \eta^N) - \psi^{N}\\
    & = \alpha^N_3(A,Z,M,Y)\{Y - \theta^N_3(A,Z,M,W)\}\\
    &+\alpha^N_2(A,Z,W)\{b^N_3(Z,M,W;\theta_3^N) - \theta^N_2(A,Z,W)\}\\
    &+\alpha^N_1(A,W)\{b^N_2(Z,W;\theta^N_2) - \theta^N_1(A,W)\}\\
    &+ b^N_1(W;\theta^N_1)-\psi^{N},
\end{align*}}%
and
{\small
\begin{align*}
\varphi^R( X; \eta^R)&=\bar\varphi^R( X; \eta^R) - \psi^R\\
& = \alpha^R_4(A,Z,M,W)\{Y- \theta_4^R(A,Z,M,W)\}\\
&+ \alpha^R_{3}(A,M,W)\{b^R_{4}(\Zp,M,W; \theta_{4}^R)- \theta^R_{3}(A,M,W)\}\\
&+ \alpha^R_{2}(A,Z,W)\{b^R_{3}(M,W; \theta_{3}^R)- \theta^R_{2}(A,Z,M)\}\\
&+ \alpha^R_{1}(A,W)\{b^R_{2}(Z,W; \theta_{2}^R)- \theta^R_{1}(A,W)\}\\
&+ b^R_{1}(W; \theta^R_{1}) - \psi^R
\end{align*}}%
satisfy the first order von Mises expansion (\ref{eq:von-mises}) for the parameters $\psi^N$ and $\psi^R$, respectively, with second-order terms $R^N_2(\eta_\F^N, \eta_\P^N)$ and $R^R_2(\eta_\F^R, \eta_\P^R)$ given in the supplementary materials. Furthermore, $\var[\varphi^{N}(X; \P)]$ and $\var[\varphi^{R}(X; \P)]$ are the corresponding non-parametric efficiency bounds in a model with observed data $X=(W, A, Z, \Zp, M, Y)$. Here we use notation $\bar\varphi^{N}(X; \eta^N)$ and $\bar\varphi^{R}(X; \eta^R)$ to introduce the uncentered gradients. 
\end{theorem}

The above von Mises expansion characterizes the non-parametric efficiency bound for estimation of $\psi^E:E\in\{N,R\}$ in at least two senses. First, according to the convolution theorem \citep{Bickel97}, the optimal asymptotic distribution for any regular estimator of $\psi^E$ is a Gaussian distribution centered at zero with variance $\var[\varphi^{E}(X; \eta_\P)]$. Second, \cite{vandervaart2002} shows that $\var[\varphi^{E}(X; \P)]$ is the local asymptotic minimax
  efficiency bound for estimation of $\psi^E$ in the sense that, for
  any estimator sequence $\psi^E_n$:
  \[\inf_{\delta
      >0}\liminf_{n\to\infty}\sup_{\Q:V(\Q-\P)<\delta}n\mathbb{E}\{\psi^E_n
    - \psi^E(\Q)\}^2\geq \var_\P[\varphi(Z;\eta_P)],\] where
  $V(\cdot)$ is the variation norm and $\mathbb{E}$ denotes expectation. We added indices $\P$
  and $\Q$ to emphasize sampling under $\P$ or $\Q$, and used notation
  $\psi^E(\Q)$ to denote the parameter computed at an arbitrary
  distribution. Furthermore, the functions $\varphi^{E}(X; \P)$ are unbiased estimating equations in the following sense, which is a corollary of Theorem~\ref{theo:vonm}.
  \begin{coro}[Robustness of estimating equation] \label{Robustness}
For $E\in\{N,R\}$ and any nuisance parameter $\tilde \eta^E$, we have $\E[\bar\varphi^{E}(X; \tilde\eta^E)]=\psi^E$ if, for each $k\in\{1,2,3,4\}$, we have either $\tilde\theta_k^E=\theta_k^E$ or $\tilde\alpha_k^E=\alpha_k^E$.
\end{coro}
According to the above corollary and efficiency discussion, the functions $\bar\varphi^E$ can be used to construct estimators with desirable properties such as double robustness, efficiency, and asymptotic linearity. We construct such estimators in the following section.
\section{Estimators and their sampling distribution}\label{sec:estima}
Having derived the corresponding von Mises expansions, in this section we focus on the development of estimation algorithms. These estimators have multiple ingredients. First, it is necessary to specify an algorithm for the construction of a plug-in estimator. This estimator will be constructed by a sequential regression procedure using the definitions of $\theta_j$ and $b_j$ given in expressions (\ref{nonprimetheta}) and (\ref{primetheta}). For instance, to construct a plug-in estimator for $\psi^N$, we first regress the outcome $Y_i$ on $(A_i,Z_i,M_i,W_i)$, using flexible data-adaptive regression procedures to obtain an estimate $\hat \theta_3^N$. We then use this estimate to compute the pseudo-outcome $b_3^N(a_1,Z_i,M_i,W_i;\hat \theta_3^N)$ by computing the predictions of $\hat \theta_3^N$ in a new dataset where $A_i$ is fixed to $a_1$. This pseudo-outcome is then regressed on $(A_i,Z_i,W_i)$ to obtain an estimate $\hat\theta_2^N$, which is then used to construct a new pseudo-outcome $b_2^N(a_2,Z_i,W_i;\hat \theta_2^N)$. This pseudo-outcome is regressed on $(A_i,W_i)$, leading to an estimate $\hat\theta_1^N$, which is used to compute $b_1^N(W_i,\hat\theta_1^N)$. The plug-in estimator for $\psi^N$ is an average across $i$ of $b_1^N(W_i,\hat\theta_1^N)$. By the analyses of the previous section, these plug-in estimators will have a first order bias of magnitude $-\E[\varphi(X, \hat\eta^N)]$. As described in the previous section, an estimate of this bias can be added back to the plug-in estimator to obtain a ``de-biased'' one-step estimator. In general, while the expectation in $\E[\varphi(X, \hat\eta^E)]$ can be estimated with an empirical average, and the parameters $\theta^E$ in $\eta^E$ can be estimated with sequential regression, it remains to develop estimators for the parameters $\alpha^E$ in expressions (\ref{nonprimealpha}) and (\ref{primealpha}). If $M$ and $Z$ are categorical, these parameters may be estimated by plugging in probability estimates obtained using off-the-shelf machine learning classification methods. However, if either of these variables is continuous or high-dimensional, plug-in estimation becomes computationally intractable, as it would require estimating conditional densities and/or integrals with respect to those densities. 

To avoid this problem, we instead adopt ideas from a recent strand of articles on \textit{Riesz learning} \citep{chernuzhukov2022riesznet,chernuzhukov2022debiased}. The foundational idea behind Riesz learning is that for an arbitrary $b$, if $\E[b(X,\theta)]$ is a continuous linear functional of $\theta$, then by the Riesz representation theorem there exists a unique function $\alpha$ such that $\E[b(X;\theta)] = \E[\alpha(X)\theta(X)]$ for all $\theta$ with $\E [\theta(X)^2] < \infty$.  In the proof of the proposition below we show that the parameters in (\ref{nonprimealpha}) and (\ref{primealpha}) correspond to Riesz representers of functionals that involve the $b$ functions in (\ref{nonprimetheta}) and (\ref{primetheta}). The mean squared error (MSE) loss function then provides an approach to estimate the $\alpha$ parameter directly, avoiding estimation of densities and integrals required for the plug in estimator. Specifically, we have
\begin{align*}
    \alpha &= \arg\min_{\tilde\alpha}\E[(\alpha(X) - \tilde\alpha(X))^2]\notag\\
    &= \arg\min_{\tilde\alpha}\E[\alpha(X)^2 - 2\tilde\alpha(X)\alpha(X)+\tilde\alpha(X)^2]\notag\\
        &= \arg\min_{\tilde\alpha}\E[\tilde\alpha(X)^2 - 2\tilde\alpha(X)\alpha(X)]\notag\\
                                &= \arg\min_{\tilde\alpha}\E[\tilde\alpha(X)^2 - 2b(X, \tilde\alpha)],
\end{align*}
where the third equality follows by ignoring the constant $\E[\alpha^2(X)]$ in the minimization problem, and the last one from the Riesz representation theorem. Thus, estimation may proceed by carrying out the minimization problem above within a given class of functions $\tilde\alpha$ (e.g., neural networks, regression trees, etc.). It is straightforward to verify that for $b_1^N$ and $b_1^R$ defined in (\ref{nonprimetheta}) and (\ref{primetheta}), the corresponding functions satisfying the Riesz representation theorem are $\alpha_1^N$ and $\alpha_1^R$ given in (\ref{nonprimealpha}) and (\ref{primealpha}). However, extra calculations are necessary to derive appropriate loss functions for the other $\alpha$ parameters. We summarize the result of those calculations in the following proposition.
\begin{proposition} \label{prop:loss_alpha}
    The MSE loss functions for $(\alpha_1^N, \alpha_2^N, \alpha_3^N)$ may be expressed as 
{\small    \begin{equation} \label{loss_alpha_nonprime}
    \begin{aligned}
        \alpha_1^N &= \arg\min_{\alpha}\E[\alpha(A, W)^2 - 2b_1^N(W, \alpha)], \\ 
        \alpha_2^N & = \arg\min_{\alpha}\E\{\alpha(A, Z, W)^2 - 2\alpha_1^N (A, W) b^N_2(Z,W, \alpha)\}, \\
        \alpha_3^N & = \arg\min_{\alpha}\E\{\alpha(A, Z, M, W)^2 - 2\alpha_2^N (A, Z, W) b^N_3(Z,M,W, \alpha)\},
    \end{aligned}
    \end{equation}}%
and the MSE loss functions for $(\alpha_1^R, \alpha_2^R, \alpha_3^R, \alpha_4^R)$ may be expressed as
{\small    \begin{equation} \label{loss_alpha_prime}
    \begin{aligned}
\alpha_{1}^R &= \arg\min_{\alpha}\E[\alpha(A,W)^2 - 2b_{1}^R(W, \alpha)],\\
\alpha_{2}^R &= \arg\min_{\alpha}\E\{\alpha(A,Z,W)^2 - 2\alpha_{1}^R(A,W)b_{2}^R(Z,W,\alpha)\},\\
\alpha_{3}^R &= \arg\min_{\alpha}\E\big\{\alpha(A,M,W)^2 - 2\alpha_{2}^R(A,Z,W)b_{3}^R(M,W;\alpha)\big\},\\
\alpha_4^R &=  \arg\min_{\alpha}\E\big\{\alpha(A,Z,M,W)^2 - 2\alpha_{3}^R(A,M,W)b_4^R(\Zp,M,W;\alpha)\big\}.
\end{aligned}
\end{equation}}%
\end{proposition}
Estimates of $\alpha_j^N$ and $\alpha_k^R$ may then be constructed by sequentially solving the above optimization problems in a given class of estimators $\alpha$. For practical reasons,  our simulations and illustrative study  will focus on deep learning since it is one of the few machine learning regression procedures with off-the-shelf software that allows specification of custom loss functions. The development of optimization methods for other function classes (e.g., ensembles of regression trees) is the subject of future work.

\begin{remark}
    Prior literature on the development of doubly-robust estimators for mediational effects used a reparameterization of the Riesz representers for estimation \citep[e.g,][]{tchetgen2013inverse,zheng2012targeted, diaz2021nonparametric}. For example, in $\alpha_2^N$, by noting that $\frac{\P (z \mid a_3, w)}{\P(z \mid a, w)} = \frac{\P (a_3 \mid z, w) \P(a \mid w)}{\P(a \mid z, w)\P(a_3 \mid w)}$, one could transform estimation of the density ratio with respect to $Z$ to estimation of density ratios with respect to the exposure $A$. However, %while similar tricks were used in many previous papers, this reparametrization does not reduce the computational complexity when the exposure is continuous. In comparison, our approach can be naturally extended to continuous treatments, see \sec\ref{sec:mtp}). Moreover, even under a simpler case when the exposure is binary, 
    in general, the assumptions required to analyze the sampling distribution of such an estimator become more difficult to satisfy. For further details, we refer the reader to Assumption 1 in \sec 4.2 of \cite{rudolph2024practical}.
\end{remark}

\subsection{Cross-fitting}
The analysis of the asymptotic properties of the estimators in \sec\ref{sec:asymp} below can be proved using certain results for empirical process under a Donsker assumption roughly stating that the estimated parameters $\hat\eta^N$ and $\hat\eta^R$ are in function classes of bounded entropy. To avoid imposing that assumption, which may not hold for some of the more flexible machine learning approaches, we use
use cross-fitting~\citep{klaassen1987consistent,zheng2011cross,
chernozhukov2016double} when constructing the estimators $\hat\eta^N$ and $\hat\eta^R$. 
 Let ${\cal V}_1, \ldots, {\cal V}_J$ denote a random partition of the
index set $\{1, \ldots, n\}$ into $J$ prediction sets of approximately
the same size. That is, ${\cal V}_j\subset \{1, \ldots, n\}$;
$\bigcup_{j=1}^J {\cal V}_j = \{1, \ldots, n\}$; and
${\cal V}_j\cap {\cal V}_{j'} = \emptyset$. In addition, for each $j$,
the associated training sample is given by
${\cal T}_j = \{1, \ldots, n\} \setminus {\cal V}_j$. Let
$\hat \eta_j^N$ and $\hat \eta_j^R$ denote the parameter estimates using training data 
$\mathcal T_j$. Letting $j(i)$ denote the index of the prediction set
which contains observation $i$, cross-fitting entails using only
observations in $\mathcal T_{j(i)}$ for fitting models when making
predictions about observation $i$. The final cross-fitted estimators of $\psi^N$ and $\psi^R$ are computed as:
\begin{equation}\label{eq:estimators}
    \hat \psi^N = \frac{1}{n}\sum_{i=1}^n\bar \varphi^N(X_i,\hat \eta_{j(i)}^N),\quad\text{and}\quad \hat \psi^R = \frac{1}{n}\sum_{i=1}^n\bar \varphi^R(X_i,\hat \eta_{j(i)}^R),
\end{equation}
respectively.
\subsection{Consistency and asymptotic sampling distribution}\label{sec:asymp}

We will first state the main result of this section, which describes the asymptotic behavior of the estimators under assumptions on the consistency of the nuisance estimators $\hat\eta^N$ and $\hat\eta^R$ and on the bounds of the parameters $\alpha^N_j$ and $\alpha^R_j$ and their estimators. The main building blocks to prove this theorem are the von Mises expansion given in Theorem~\ref{theo:vonm}, as well as tail inequalities applied to the corresponding cross-fitted empirical processes \citep[see][for a review of the main techniques]{kennedy2022semiparametric}. First, we introduce the following assumption, which is akin to the overlap or positivity assumption invoked for estimation of the average treatment effect.
\begin{assumption}[Bounded Riesz representers]\label{ass:pos}
    For $E\in\{N,R\}$, assume the estimated functions $\hat\alpha^E_j$ are bounded in the sense that there exists a constant $M<\infty$ such that $\P(\hat\alpha^E_j(X)\leq M)=\P(\alpha^E_j(X)\leq M)=1$,
\end{assumption}
The statement of the results in this section will benefit from explicitly describing the form of the second order term in the von Mises expansion. To that end, for each estimator $\hat\theta_{k,j(i)}^N$, we define the following parameters:
{\small
\begin{align*}
    \tilde \theta_{2,j(i)}^N(a,z,w)&=\E[b_3^N(Z,M,W;\hat \theta_{3,j(i)}^N)\mid A=a, Z=z, W=w];\\
        \tilde \theta_{1,j(i)}^N(a,w) &= \E[b_2^N(Z,W;\hat \theta_{2,j(i)}^N)\mid A=a,W=w],
\end{align*}}%
where we note that the expectations are taken with respect to the distribution of $(W,Z,M)$ with the estimators $\hat\theta^N_{3,j(i)}$ and $\hat\theta^N_{2,j(i)}$ fixed. Define
{\small
\begin{align*}
    R_2^N(\hat\eta_{j(i)}^N,\eta^N) &= \sum_{k=1}^3\E\big[\{\hat\alpha_{k,j(i)}^N(X) - \alpha_k^N(X)\}\{\hat \theta_{k,j(i)}^N(X)-\tilde\theta_{k}^N(X)\}\big],\\
        R_2^R(\hat\eta_{j(i)}^R,\eta^R) &= \sum_{k=1}^4\E\big[\{\hat\alpha_{k,j(i)}^R(X) - \alpha_k^R(X)\}\{\hat \theta_{k,j(i)}^R(X)-\tilde\theta_{k}^R(X)\}\big],
\end{align*}}%
with $\tilde \theta_{k,j(i)}^R$ defined analogously, $\tilde \theta_{3,j(i)}^N= \theta_3^N$, and $\tilde \theta_{4,j(i)}^R= \theta_4^R$. We will use the following assumption.
\begin{assumption}[Rate of convergence of second-order term]\label{ass:rate}
    For $E\in\{N,R\}$, assume $R_2^E(\hat\eta_{j(i)}^E,\eta^E)=o_\P(n^{-1/2})$.
\end{assumption}
The following theorem is the main result of this section.
\begin{theorem}[Asymptotic linearity and weak convergence] \label{theo:asymnormal} 
Assume \ref{ass:pos}. Then we have
\[\hat \psi^E - \psi^E 
= \frac{1}{n} \sum_{i=1}^n\{\varphi^E(X_i; \eta^E) + R_2^E(\hat\eta^E_{j(i)},\eta^E)\}+o_{\P}(n^{-1/2}).\]
Under the additional assumption \ref{ass:rate}, the central limit theorem yields $\sqrt{n}(\hat \psi^E- \psi^E) \rightsquigarrow N(0,\sigma^{2,E})$, where $\sigma^{2,E}=\var\{ \varphi^E(X; \eta^E)\}$ is the non parametric efficiency bound in a model with observed data $(W, A, Z, \Zp, M, Y)$.
\end{theorem}

The above theorem has two important implications. First, it allows us to obtain estimates of the uncertainty around the estimates in the form of confidence intervals and standard errors. For instance, an estimate $\hat\sigma^{2,E}$ may be computed as the empirical variance of $\varphi^E(X_i;\hat\eta_{j(i)}^E)$, and a $(1-\alpha)100\%$ Wald-type confidence interval may be computed as $\hat\psi^E\pm z_{1-\alpha/2}\times(\hat\sigma^{2,E}/n)^{-1/2}$. Likewise, this theorem provides the basis to use the Delta method to understand the asymptotic distribution of the mediational parameters  defined in Proposition \ref{prop:equivalence}, as well as other important contrasts such as the proportion mediated or the proportion of the effect operating through a specific path. 

The above theorem requires that all the nuisance parameters in $\eta^E$ are estimated consistently at the right rates in order to achieve $\sqrt{n}$-consistency and asymptotic normality of $\hat\psi^E$. A weaker result, stating merely the assumptions required for consistency of the estimator, is presented in the following corollary.  

\begin{coro}[Double robustness] \label{Robustness}
For a function $f$, let $||f||^2=\int f^2(x)\dd\P(x)$. For $E\in\{N,R\}$ and each $k\in\{1,2,3,4\}$, assume either $||\hat\theta^E_{k,j(i)} - \tilde\theta^E_k||=o_\P(1)$ or $||\hat\alpha^E_{k,j(i)} - \alpha^E_k||=o_\P(1)$. Then we have $\hat\psi^E=\psi^E+o_P(1)$.
\end{coro}

The above corollary is a consequence of Theorem~\ref{theo:asymnormal} and the Cauchy-Schwarz inequality applied to the definition of $R_2^E$. An important consequence is that the estimators are consistent if, for each of the terms involved in the canonical gradients in Theorem~\ref{theo:vonm}, either the Riesz representer or the outcome regressions are estimated consistently. It is important to note that the results of this section are given in terms of assumptions that require consistency of $\hat\theta_k^E$ as an estimator of $\tilde\theta_k^E$, i.e., only the latest expectation in the sequential regression procedure is required to be estimated consistently. This has  connections to results in the literature for estimation of effects in longitudinal data models \citep{luedtke2017sequential, rotnitzky2017multiply}, where it is shown that certain one-step estimators cannot achieve sequential double robustness as in Corollary~\ref{Robustness} above. The reason our estimators allow for sequential double robustness is that, in contrast to \cite{luedtke2017sequential} and \cite{rotnitzky2017multiply}, we estimate the Riesz representer directly instead of using plug-in estimators. On the other hand, according to Proposition \ref{prop:loss_alpha} our approach to estimate the Riesz representers is sequential, so that consistent estimation of $\alpha_k^E$ requires consistent estimation of $\alpha_{k-1}^E$.

\subsection{Practical construction of $\Zp$}\label{sec:genzp}% and implications on asymptotic results}

Recall that our estimators assumed the existence of a variable $\Zp$ such that (\ref{eq:defZp}) holds. To introduce our main approach to construct this variable, consider a hypothetical case in which $(A,W)$ are discrete. Then, it can be seen that a simple permutation of $Z$ within strata of $(A,W)$ would satisfy (\ref{eq:defZp}). To construct $\Zp$ in the general case, we will conduct a permutation of the original $Z$ variable as follows. Define ${\bf Z} = (Z_1,\ldots,Z_n)$ to be an $n\times p$ matrix, and define $\bf \Zp$, $\bf A$, and $\bf W$ accordingly. For two matrices $\bf C$ and $\bf D$ of sizes $n\times p_1$ and $n\times p_2$, respectively, we let $(\bf C, D)$ denote a column-wise concatenation of size $n\times (p_1+p_2)$.
The permutation $\bf \Zp$ is obtained by applying a permutation matrix, denoted with $\bf \Pi$, to the original data $\bf Z$, i.e., we  define ${\bf \Zp} = {\bf \Pi Z}$. To ensure that $\Zp$ and $Z$ have the same distribution conditional on $(A, W)$, ideally we would choose a permutation matrix such that $({\bf A}, {\bf W})=\bf \Pi({\bf A}, {\bf W})$. With continuous data $(A,W)$, it will be impractical to find such a permutation matrix. Thus, we focus on finding a permutation matrix that minimizes a suitably defined distance between $({\bf A}, {\bf W})$ and $\bf \Pi({\bf A}, {\bf W})$, so that $({\bf A}, {\bf W}) \simeq \bf\Pi({\bf A}, {\bf W})$. To compute this matrix $\bf\Pi$, we follow the approach presented in \cite{doran2014permutation}. This approach relies on a distance function $d$ and a matrix $\bf D$ with $(i,j)$-th element $D_{i,j}=d((A_i, W_i), (A_j,W_i))$. Specifically, we solve
\[
\min_{\bf\Pi \in \mathcal P} \sum_id\{(A_i,W_i), ({\bf\Pi}({\bf A}, {\bf W}))_i\} = \min_{{\bf\Pi} \in \mathcal P} \sum_{i,j} \Pi_{ij} D_{ij} = \min_{\bf\Pi \in \mathcal P} \operatorname{Tr}(\bf\Pi D),
\]
where $\mathcal P$ is the collection of doubly stochastic matrices (matrices whose rows and columns both sum to one) with zero trace, which are linear constraints of this optimization problem. The zero-trace condition is enforced to make sure that $\bf\Pi$ will never be an identity matrix. Note that the first equality holds because ${\bf\Pi}_{i,j}=1$ if and only if $(A_i,W_i)$ is permuted to $(A_j,W_j)$. We solve this optimization problem using the simplex algorithm \citep{6539ce83c8db467ead88d4b6bc7f7206}.

%  With a slight abuse of notations, $Z^*$ is obtained by generating a permutation matrix (called $\Pi$), which can break the dependence between $Z$ and $M$, and then applying $P$ on the original data $Z$ (namely we define $Z^* = PZ$). Moreover, to ensure that $Z^*$ and $Z$ follow the same distribution conditional on $A, W$, ideally we want $(A, W)$ to be unchanged after applying $P$ on it. Practically, we want the distance between $(A, W)$ and $P(A, W) = (PA, PW)$ to be minimized, so that $(A, W) \simeq P(A, W)$. To get this $P$, we follow the approach defined in \cite{doran2014permutation}, which means we compute the distance matrix of dataset $(A, W)$ (where we define as $D$ with $D_{ij} = d((a_i, w_i), (a_j, w_j))$, and solve the following constrained optimization problem
% \[
% \min_{P \in \mathcal P} d((A, W), P(A, W)) = \min_{P \in \mathcal P} \sum_{ij} P_{ij} D_{ij} = \min_{P \in \mathcal P} \operatorname{Tr}(PD),
% \]
% where $\mathcal P$ is the collection of doubly stochastic matrices (matrices whose rows and columns both sum to one) with zero trace, which are linear constraints of this optimization problem. Note that the first equality holds because the feasible region for this optimization problem is a convex hull of permutation matrices, and the minimal for this problem is one of the vertexes, which is a permutation matrix. In the case when $P$ is a permutation matrix, $d((A, W), (PA, PW)) = \sum_{ij} P_{ij} D_{ij}$. 
% The zero-trace condition is enforced to make sure that $P$ will never be an identity matrix. 

% Describe how to construct it

\section{Extension to non-binary treatments using modified treatment policies}\label{sec:mtp}

%{\color{red} RL: As there will always be people confusing MTP and dynamic treatment regime, should we make a distinction here, or just blur it?}
The developments in the previous sections assumed that the exposure variable $A$ is binary, and that interest lies in decomposing effects defined as a contrast between outcome expectations in a hypothetical world where $\P(A=1)=1$ vs. a hypothetical world where $\P(A=0)=1$. In this section we focus on an extension of the methods to problems where the exposure is non-binary, and where the focus is on decomposing effects other than the average treatment effect. In particular, we focus on the methodology of modified treatment policies \citep[MTP,][]{Diaz12, Haneuse2013, diaz2023lmtp}. MTPs have also been called \textit{dynamic interventions the depend on the natural value of treatment} \citep{robins2004effects, young2014identification, richardson2013single}. Let $d_1(a,w)$ and $d_0(a,w)$ denote two functions that map a treatment value $a$ amd a covariate value $w$ into potential post-intervention treatment values. Causal effects of modified treatment policies are defined as contrasts $\E[Y(d_1) - Y(d_0)]$, where $Y(d)$ denotes the counterfactual outcome that would have been observed in a hypothetical world where treatment had been assigned according to $d(A, W)$, where we note that the MTP definition recovers the average treatment effect of a binary exposure by setting $d_1=1$ and $d_0=0$. Modified treatment policies allow the user to define interesting causal contrasts for continuous or multivariate exposures by careful consideration of the functions $d$ being contrasted. For instance, a researcher interested in the effect of an increase of 10\% in a continuous exposure might consider setting 
\[d_1(a,w) = 
\begin{cases}
1.1\times a &\text{ if }1.1\times a < u(w)\\
a &\text{ otherwise, }
\end{cases}\]
and $d_0(A,W)=A$, where $u(w)$ is largest feasible value of $A$ within strata  $W=w$. For a discussion on the flexibility of the MTP framework and other interesting policy definitions we refer the reader to the original articles as well as the review by \cite{hoffman2023introducing}.% and \cite{williams2023lmtp} {\color{red} RL: I remember one of the reviewers pointed out the incorrectness in \cite{williams2023lmtp}?}.

Mediation analysis leveraging modified treatment policies was recently discussed by \cite{gilbert2024identification}. In that work, the authors develop estimators for general longitudinal settings, but the estimation methods are restricted to categorical mediators. Furthermore, because the methods are developed within the randomized interventional framework, they are subject to the critique that they do not measure mechanisms in special settings, as discussed by \cite{miles2023causal}.  In Appendix~\ref{sec:genrt} we present a generalization of the recanting twin effects of \cite{diaz2024non} and \cite{vo2024recanting}  to decompose the effect of modified treatment policies as defined above. Although the details of the definition and identification of our generalization are important, we relegate their discussion to the appendix so that we can focus on a discussion of the main point of our paper, which is the development of general estimation techniques. 

Because the definitions and results of this section generalize those of prior sections, in a slight abuse of notation we will use the same symbols to denote the objects that will be generalized. This will allow us to avoid re-stating theorems and results that hold true with the updated definitions presented in this section. For values $(i,j,k,l) \in \{0,1\}^4$, we define $A^s = d_s(A, W)$, and let

{\footnotesize
\begin{equation}
\begin{aligned}
    \psi^R(i,j,k,l)  & = \E\Bigg[\E\bigg[\E\Big[\E \big[ \E (Y \,|\, A = A^i, \Zp, M, W ) \,\big|\, A=A^j,M, W\big ]\,\Big|\, A=A^k, Z, W\Big]\,\bigg|\, A=A^l, W\bigg]\Bigg],\\
    \psi^N(j,k,l)   & = \E \bigg[\E \Big[ \E \big[ \E(Y\mid A=A^{j}, Z, M, W) \,\big|\, A = A^{k}, Z, W\big] \,\Big|\, A = A^{l}, W\Big]\bigg].
\end{aligned}\label{eq:defpsimtp}    
\end{equation}}%

Similar to Proposition~\ref{prop:equivalence}, we have the following result establishing the generality of the above parameters $\psi^N$ and $\psi^R$.
\begin{proposition}[Identification of parameters for mediation using modified treatment policies]\label{prop:equivalencemtp}
The identification formulas for the causal parameters defined by \cite{gilbert2024identification} under a single time-point can be written in terms of $\psi^N$ and $\psi^R$ as:

{\footnotesize
\begin{align*}
        \operatorname{RIDE} &= \psi^R(1,1,0,0) - \psi^R(0,0,0,0), &\operatorname{RIIE} &= \psi^R(1,1,1,1) - \psi^R(1,1,0,0),
\end{align*}}%, 
and the identification formulas for the path-specific recanting twin effects defined in Theorem~\ref{theo:idenmtp} of Appendix~\ref{sec:genrt} can be written as 
{\footnotesize
\begin{align*}
    \operatorname{RT}_1 &= \psi^N(1,1,1)-\psi^N(0,1,1), &\operatorname{RT}_2 &= \psi^R(0,1,1,1) - \psi^R(0,0,1,1),\\
    \operatorname{RT}_3 &= \psi^R(0,0,1,1) - \psi^R(0,0,1,0), &\operatorname{RT}_4 &= \psi^N(0,1,0) - \psi^N(0,0,0).
\end{align*}}%
\end{proposition}

Likewise, the formula for the von Mises expansion of mediational parameters based on modified treatment policies is identical to that of mediational for the average treatment given in Theorem~\ref{theo:vonm}, with the following updated definitions of $\alpha_k^E$ and $\theta_k^E$. For $\psi^N$,  define

{\footnotesize
\begin{equation} \label{nonprimethetamtp}
\begin{aligned}
\theta^N_3(a,z,m,w)&=\E(Y\mid A=a,Z=z,M=m,W=w), &     b^N_3(a,z,m,w;\theta_3^N)&=\theta^N_3(d_{j}(a, w),z,m,w),\\
    \theta^N_2(a,z,w)&=\E[b^N_3(A,Z,M,W;\theta_3^N)\mid A=a, Z=z, W=w],    & b^N_2(a,z,w;\theta_2^N)&=\theta^N_2(d_{k}(a,w),z,w),\\
        \theta^N_1(a,w) &= \E[b^N_2(A,Z,W;\theta_2^N)\mid A=a,W=w],     & b^N_1(a,w;\theta_1^N)&=\theta^N_1(d_{l}(a,w),w),\\
\end{aligned}
\end{equation}}%
as well as
{\footnotesize
\begin{equation}
\begin{aligned} \label{nonprimealphamtp}
         \alpha^N_1(a,w) & = \frac{\P_{d_l}(a\mid w)}{\P(a\mid w)},\\
        \alpha^N_2(a,z,w) & = \frac{\P_{d_k}(a\mid w)}{\P(a\mid w)}\frac{\P(z\mid A=d_l(a,w), w)}{\P(z\mid a, w)},\\
        \alpha^N_3(a,z,m,w) & = \frac{\P_{d_j}(a\mid w)}{\P(a\mid w)}\frac{\P(m\mid A=d_k(a,w), z, w)}{\P(m\mid a, z, w)}\frac{\P(z\mid A=d_l(a,w), w)}{\P(z\mid a, w)}.
\end{aligned}
\end{equation}}%
For $\psi^R$, define
{\footnotesize
\begin{equation} \label{primethetamtp}
\begin{aligned}
\theta^R_4(a,z,m,w)&=\E(Y\mid A=a,Z=z,M=m,W=w), &b^R_4(a,z,m,w;\theta_4^R)&=\theta^R_4(d_i(a,w),z,m,w),\\
\theta^R_{3}(a,m,w)&=\E[b^R_4(A,\Zp,M,W;\theta_4^R)\mid A=a, M=m, W=w], &b^R_{3}(a,m,w;\theta_{3}^R)&=\theta^R_{3}(d_j(a, w),m,w), \\
\theta^R_{2}(a,z,w)&=\E[b^R_{3}(A,M,W;\theta_{3}^R)\mid A=a, Z=z, W=w], & b^R_{2}(a,z,w;\theta_{2}^R)&=\theta^R_{2}(d_k(a,w),z,w), \\
\theta^R_{1}(a,w) &= \E[b^R_{2}(A,Z,W;\theta_{2}^R)\mid A=a,W=w], & b^R_{1}(a,w;\theta_{1}^R)&=\theta^R_{1}(d_l(a,w),w),  \\
\end{aligned}
\end{equation}}%
as well as
{\footnotesize
\begin{equation} \label{primealphamtp}
\begin{aligned}
\alpha_{1}^R(a,w) & = \frac{\P_{d_l}(a\mid w)}{\P(a\mid w)};\\
\alpha_{2}^R(a,z,w) & = \frac{\P_{d_k}(a\mid w)}{\P(a\mid w)}\frac{\P(z\mid A=d_l(a,w), w)}{\P(z\mid a, w)};\\
\alpha_{3}^R(a,m,w) & = \frac{\P_{d_j}(a\mid w)}{\P(a\mid w)}\frac{\int_z \P(m\mid A=d_k(a,w), z, w) \dd \P(z\mid A=d_l(a,w),w) }{\P(m\mid a, w)};\\
\alpha_4^R(a,z,m,w) & = \frac{\P_{d_i}(a\mid w)}{\P(a\mid w)}\frac{\P(z\mid A=d_j(a,w), w)}{\P(z\mid a, w)}\frac{\int_z \P(m\mid A=d_k(a,w), z, w) \dd \P(z\mid A=d_l(a,w),w) }{\P(m\mid a, z, w)}.
\end{aligned}
\end{equation}}%
With these generalized definitions of the nuisance parameters, Proposition~\ref{prop:loss_alpha} may  be used to estimate the Riesz representers, expression (\ref{eq:estimators}) provides a generalized formula for constructing the estimators, and Theorem~\ref{theo:asymnormal} provides the basis for constructing confidence intervals and hypothesis tests. 

\section{Numerical studies}\label{sec:simula}
We conduct a Monte Carlo simulation study to illustrate the performance of our proposed estimation algorithm. We evaluate the estimators with the data generated by the following data generating process (DGP):

\begin{table}[H]
\setlength{\tabcolsep}{1pt}
\centering
\small
\label{tab:dgm}
\begin{tabular}{rl}
%\toprule
$W_1, W_2, W_3$ & $\sim Be(2, 3)$ \\ 
$A | W $ & $ \sim Bern (\operatorname{logit}^{-1}(0.5 W_1 + 0.5 W_2 - 1))$ \\
$Z_1 | A, W$ & $ \sim TN (-1, 1, -0.4 + \epsilon A + 0.2 W_3^2, 1)$ \\
$Z_2 | Z_1, A, W$ & $ \sim TN (-1, 1, 0.2 - \epsilon A + 0.5 \sin (W_2), 1)$ \\
$M_1 | Z_1, Z_2, A, W$ & $ \sim TN (-1, 1, -0.5 + \lambda_1 Z_1 + \lambda_2 A + 0.4 W_2 + 0.2 W_3, 1)$ \\
$M_2 | M_1, Z_1, Z_2, A, W$ & $ \sim TN (-1, 1, -0.5 + \lambda_1 Z_2 + \lambda_2 A + 0.4 W_1 + 0.2 W_3, 1)$ \\
$Y | M_1, M_2, Z_1, Z_2, A, W$ & $ \sim N (0.2 M_1 + 0.2 M_2 + \gamma_1 Z_1 / 2 + \gamma_1 Z_2 / 2 + \gamma_2 A - 0.5 \cos (W_1) - 1.5, 1)$. \\
\end{tabular}
\label{tab:sim}
\end{table}
Here, $W = (W_1, W_2, W_3)^\top$ is a three-dimensional random vector. \\ $(\epsilon, \lambda_1, \lambda_2, \gamma_1, \gamma_2)^\top = (0.5, 0.4, 0.6, 0.6, 0.4)^\top $ are pre-specified parameters; $Be(\alpha, \beta)$ denotes a Beta distribution with shape parameters $\alpha$ and $\beta$; $Bern(\pi)$ denotes a Bernoulli distribution with probability $\pi$; $N(\mu, \sigma^2)$ denotes a normal distribution with mean $\mu$ and variance $\sigma^2$; $TN(a, b, \mu, \sigma^2)$ denotes a truncated normal distribution derived from $N(\mu, \sigma^2)$, but with values in $[a, b]$. 

We evaluate estimators of the natural direct and indirect effects (NDE/NIE), interventional effects (IDE/IIE), and recanting-twins effects. We note that, in addition to the 4 path-specific effects, the methodology based on recanting twins also provides a fifth parameter, which quantifies the amount of intermediate confounding by $Z$, and can therefore be used to test the null hypothesis that the estimates can be interpreted as natural path-specific effects \citep[see][]{vo2024recanting}. Thus, our simulation has a total of nine parameters of interest. For natural effects, we let $\epsilon = \lambda_1 = 0$ to make the effects identifiable. Bias, $\sqrt n$-bias, $n$MSE and coverage of the $95\%$ confidence interval of these parameters are used for our evaluation. 

For estimating the Riesz representers $\alpha$ we use deep learning \citep{lecun2015deep} using the loss functions in (\ref{loss_alpha_nonprime}) and (\ref{loss_alpha_prime}). For the sequential regression function $\theta$, we use an ensemble of LightGBM \citep{ke2017lightgbm}, mean function, Multivariate Adaptive Regression Splines \citep{Friedman1991} and neural networks, based on the Super Learner algorithm \citep{vanderLaanPolleyHubbard07}. 
The simulation study is conducted by drawing 500 datasets of sizes $n \in \{500, 1000, 2000\}$. The R code for implementation is publicly available on \href{https://github.com/CI-NYC/Semi-automated-efficient-learning-in-mediation-analysis/tree/main}{\texttt{Github}}. We note that this simulation setup is representative of typical practice in statistics, and that we have no reason to expect that both the Riesz representers and the sequential regressions estimators are consistent. Therefore, the properties of Theorem~\ref{theo:asymnormal} are not expected to hold exactly. 

Table \ref{tab:sim} shows the summary of the results. We observe that in most cases, the empirical coverage is close to the nominal value of $95\%$. Although the $\sqrt n$-bias does not decrease as $n$ increases, the $n$MSE decreases consistently across all causal parameters. This observation illustrates that the estimators may be appropriately trading off bias and variance. We also note that the coverage for the effect through the path $A \to Z \to Y$ seems to be decreasing as sample size increases, possibly due to the violation of the assumptions underlying the theorems that establish asymptotic normality. 

\begin{table}[h]
\caption{Simulation results for our method. }
\centering
\footnotesize
\begin{tabular}[]{ccccccccccc}
\toprule
\multicolumn{2}{c}{} & \multicolumn{5}{c}{recanting twins} & \multicolumn{2}{c}{interventional} & \multicolumn{2}{c}{natural} \\
\cmidrule(r){3-7} \cmidrule(r){8-9} \cmidrule(r){10-11} 

$n$ & metrics & AY & AZY & AZMY & AMY & Int. Confounder & IDE & IIE & NDE & NIE \\
\midrule
\multirow{4}{*}{500} & Bias & 0.016 & -0.014 & -0.020 & 0.004 & -0.003 & -0.005  & 0.005 & 0.000 & -0.005 \\
 & $\sqrt n$-Bias & 0.386 & -0.302 & -0.439 & 0.092 & -0.058 & -0.102 & 0.115 & 0.000 & -0.121\\
 & $n$MSE & 9.009 & 15.79 & 3.500 & 4.917 & 9.948 & 5.135 & 11.53 & 7.189 & 3.143\\
 & Coverage & 0.942 & 0.940 & 0.952 & 0.950 & 0.960 & 0.942& 0.960 & 0.954 & 0.940\\
 \midrule
\multirow{4}{*}{1000} & Bias & 0.023 & -0.033 & -0.015 & 0.000 & 0.003 & -0.005 & 0.002 & 0.005 & -0.004\\
 & $\sqrt n$-Bias & 0.725 & -1.032 & -0.464 & -0.002 & 0.103
 & -0.158 & 0.052 & 0.166 & -0.128\\
 & $n$MSE & 8.961 & 10.82 & 2.333 & 3.270 & 5.722 & 4.416 & 10.03 & 5.996 & 1.306\\
 & Coverage & 0.932 & 0.930 & 0.942 & 0.946 & 0.970 & 0.950 & 0.948 & 0.962 & 0.942\\
 \midrule
\multirow{4}{*}{2000} & Bias & 0.021 & -0.026 & -0.014 & -0.002 & 0.005 & 0.000 & 0.000 & 0.003 & 0.000\\
 & $\sqrt n$-Bias & 0.951 & -1.155 & -0.628 & -0.077 & 0.240 & 0.014 & 0.016 & 0.142 & 0.007\\
 & $n$MSE & 7.709 & 4.760 & 1.513 & 2.276 & 4.670 & 3.597 & 10.123 & 5.972 & 1.029\\
 & Coverage & 0.948 & 0.888 & 0.934 & 0.958 & 0.964 & 0.936 & 0.940 & 0.956 & 0.966\\
\bottomrule
\end{tabular}
\label{tab:sim}
\end{table}

\section{Illustrative Examples}\label{sec:ilustra}

We applied our proposed estimator in an effort to understand the extent to which pain management practices mediate the total effect of having a chronic pain disorder on opioid use disorder (OUD) risk by the end of 24 months of follow up among adult Medicaid beneficiaries. In this example, the exposure of interest $A$ is presence of a chronic pain condition (N=67,438) versus the absence of a chronic pain condition or physical disability (N=1,704,454)  during the first 6 months of Medicaid enrollment. We considered a bundle of 11 pain management practices $M$, all measured during the subsequent 6 months: i) opioid dose, ii) proportion of days covered by an opioid prescription, iii) number of unique opioid prescribers, iv) presence of opioid dose tapering, v) opioid-benzodiazepine coprescribing, vi) opioid-gabapentinoid coprescribing, vii) opioid-muscle relaxant coprescribing, viii) opioid-stimulant coprescribing, ix) presence of a nonopioid pain prescription, x) physical therapy, and xi) presence of a pain treatment claim not captured in the previous mediators. The outcome was new-episode (i.e., incident) opioid use disorder diagnosis. We controlled for baseline covariates related the patient's demographics (age, sex, race/ethnicity, primary language, married or partnered, household size, veteran status, and receipt of various public benefits) and mental health diagnoses (bipolar, depression, anxiety, attention deficient disorder, and other psychiatric condition), and post-exposure confounders $Z$ of a new diagnosis of depression or anxiety and mental health counseling. Information on the data and particular cohort of individuals we consider has been published previously \citep{rudolph2024pain}.

We estimate (an average treatment effect) that having a chronic pain condition increases risk of OUD over 18 months of follow-up by 0.0131 (95\% CI: 0.0108, 0.0155). This is similar to the risk difference estimated previously \citep{hoffman2024independent}. Using the proposed estimator to estimate the recanting-twins effects (Table \ref{tab:app}), we see that the indirect effect path $A \rightarrow M \rightarrow Y$ is responsible for nearly all of the total effect (80\%), contributing 0.0106 to the average treatment effect (95\% CI: 0.0082, 0.0131). The other potential indirect effect path $A \rightarrow Z \rightarrow M \rightarrow Y$ has a point estimate of zero, indicating that it does not contribute to the indirect effect. The direct effect path-specific estimates are very small, contributing little to the total effect. The parameter labeled ``Intermediate confounding'' in the table is discussed in \cite{vo2024recanting}; it is a parameter that allows us to test the hypothesis of no-intermediate confounding. If this parameter is not significantly different from zero, we can reject the hypothesis of intermediate confounding by $Z$, and therefore potentially interpret the estimates as natural path-specific effects \citep{avin2005identifiability}. We estimate this parameter to be %ile an order of magnitude smaller than the indirect effect, to nonetheless be 
 statistically significant: 0.0019 (95\% CI: 0.0016, 0.0021). This provides empirical support for the influence of intermediate confounders and highlights that natural effects would not be identifiable. 
 
 For contrast, we also estimate randomized interventional effects in Table \ref{tab:app}. The interventional total effect  of 0.0232 is much larger than the ATE. This confirms that intermediate confounding cannot be ignored and provides additional evidence of the perils of randomized interventional effects for addressing intermediate confounding \citep{miles2023causal}. The IIE contributes to 68\% of the randomized total effect, which is similar percentage as estimated for the recanting-twins effects, but the IDE is much larger than the sum of the path-specific recanting twin effects through $A\to Z$ and $A\to Z\to Y$, which highlights drawbacks of the IIE/IDE that are well-known in the literature. Furthermore, the recanting twin effects provide specific information about the path involved in the effect through $M$, specifiying that all the effect goes through $A\to M\to Y$ rather than $A\to Z\to M\to Y$. The randomized interventional effects do not provide any information of this type, instead considering both pathways as a bundle.  

\begin{table}[h]
\caption{Illustrative application results. }
\centering
\small
\begin{tabular}[]{l c c}
\toprule
\label{tab:app}
& Estimate & 95\% CI\\
\hline
\multicolumn{3}{c}{Recanting twins}\\
$A \rightarrow Y$& -3e-04& (-5e-04, 0)\\
$A \rightarrow Z \rightarrow Y$ & 9e-04&  (7e-04, 0.001)\\
$A \rightarrow Z \rightarrow M \rightarrow Y$ & 0 &(-1e-04, 1e-04)\\
$A \rightarrow M \rightarrow Y$ & 0.0106 & (0.0082, 0.0131)\\
Intermediate confounding &0.0019 & (0.0016, 0.0021)\\
\midrule
\multicolumn{3}{c}{Interventional}\\
RIDE&  0.0072& (0.0047, 0.0096)\\
RIIE& 0.0155& (0.0151, 0.0076)\\
\bottomrule
\end{tabular}
\end{table}

\section{Discussion} \label{sec:discussion}
There has been important progress in the non-parametric definition and identification of causal parameters in mediation analysis. These parameters remain underutilized in practice because estimators for many approaches remain underdeveloped. For instance, few or no methods can handle  mediators or intermediate confounders that are continuous or multivariate, a situation that is pervasive in contemporary research. 

In this manuscript, we made several contributions that will increase the utilization of these parameters in practice. First, we developed non-parametric double machine learning estimators for the six mediation parameters listed in Table \ref{tab:params}, and proved that the estimators enjoy desirable statistical properties including asymptotic linearity, weak convergence and double robustness under assumptions. To do so, we unified the identification formulas of these parameters as two statistical estimands, and then developed non-parametric estimators of these estimands with the aid of the semi-parametric efficiency theory. Our estimators are based on the EIF, which characterizes the first-order bias of plug-in estimators and non-parametric efficiency bound. The key challenge in mediation analysis with continuous or high-dimensional mediators is that estimation of the EIF requires estimation of densities of the mediator as well as integrals with respect to such densities. We address challenges in the estimation of multivariate or high-dimensional densities on the mediator by estimating $\alpha_j^E, j \in \{1, 2, 3, 4\}, E \in \{N, R\}$ directly through the use of so-called Riesz learning, without the necessity of knowing their forms. We address integration with respect to the densities of the mediator by the unification, parameterizing the integrals as conditional expectations using stratified random permutations ($Z^\pi$) of the observed variable $Z$, which we discussed in detail in \sec\ref{sec:genzp}. 

Second, we proposed estimation methods that can handle continuous or multivariate treatments, another pervasive situation for which methods remained underdeveloped. Third, we developed a publicly available R package \texttt{crumble}. The success in developing a general estimation framework for all six mediational parameters allows \texttt{crumble} to be general, where researchers can choose their parameters of interest based on their study questions without worrying about data complexity. %We believe this manuscript has the potential to unleash the power of mediation analysis and \texttt{crumble} is a useful resource to many practitioners due to its generalizability. 

There are at least two possible future directions for this work. First, for practical reasons, we used deep learning to solve the sequential optimization problems in Proposition \ref{prop:loss_alpha}, since it is one of the few machine learning regression procedures with off-the-shelf software that allows specification of custom loss functions. The development of other optimization methods for Proposition \ref{prop:loss_alpha}, such as ensembles of regression trees or gradient boosting \citep{lee2025rieszboost}, is subject to future work. Second, we note that two of the six mediational parameters quantified the path-specific effects. However, in the case of multiple mediators (e.g. two mediators $M_1, M_2$), the path-specific effects can only be quantified by treating the mediators as a bundle. The definition, identification and non-parametric estimation of path-specific effects in the case of multiple mediators is another interesting future direction. 
\section*{Acknowledgements}

Iv\'an D\'iaz and Kara Rudolph were supported through a Patient-Centered Outcomes Research Institute (PCORI) Project Program Funding Award (ME-2021C2-23636-IC) and through the National Institute on Drug Abuse (R01DA053243).

\bibliographystyle{plainnat}
\bibliography{refs}

\newpage 
\appendix

\section{Generalization of recanting twins to modified treatment policies}\label{sec:genrt}
Let $d_1(a,w)$ and $d_0(a,w)$ denote two functions that map a treatment value $a$ and a covariate value $w$ into potential post-intervention treatment values. We define causal effects based on a structural causal model defined as follows. 
\begin{definition} \label{def:scm1} Let
  $\mathcal{M}=\langle U, X, f, \mathcal{P}\rangle$ be a SCM, where
  $X=(W,A,Z,M,Y)$ are the endogenous variables,
  $U = (U_W, U_A, U_Z, U_M, U_Y) \sim \P_U$ are the exogenous
  variables, $\mathcal{P}$ is the set of allowed distributions $\P_U$,
  and $f = (f_W, f_A, f_Z, f_M, f_Y)$ are deterministic functions such
  that:
    \begin{gather*}
        W= f_W(U_W); \qquad  A = f_A(W, U_A); \qquad Z = f_Z(W, A, U_Z);\\
        M = f_M(W, A, Z, U_M); \qquad  Y = f_Y(W, A, Z, M, U_Y). 
    \end{gather*}
\end{definition}Causal effects of modified treatment policies are defined as contrasts
\[\psi=\E[Y(d_1) - Y(d_0)],\]
where $Y(d) = Y(d, Z(d), M(d, Z(d)))$, where $Y(a, z, m) = f_Y(W, a,z,m,U_Y)$ and $Z(d)$ and $M(d)$ are defined analogously. Define the paths $P_1: A\rightarrow Y$;
$P_2: A \rightarrow Z \rightarrow Y$;
$P_3: A \rightarrow Z \rightarrow M \rightarrow Y$ and
$P_4: A \rightarrow M \rightarrow Y$. Path specific effects for $P_j$ could be defined as expectations $\E[Y_{S_j} - Y_{S_{j-1}}]$ using the following counterfactuals \begin{align*}
  Y_{S_0}&=Y(d_1, Z(d_1), M(d_1, Z(d_1))),\\
  Y_{S_1}&=Y(d_0, Z(d_1), M(d_1, Z(d_1))),\\
  Y_{S_2}&= Y(d_0, Z(d_0), M(d_1, Z(d_1))),\\
  Y_{S_3}&=Y(d_0, Z(d_0), M(d_1, Z(d_0))),\\
  Y_{S_4}&=Y(d_0, Z(d_0), M(d_0, Z(d_0))).
  \end{align*}
  However, due to the recanting twins problem \citep{avin2005identifiability}, the distribution of $Y_{S_2}$ is unidentifiable. We instead propose alternative effects based on the following definition.
\begin{definition}[Recanting twins] Define $T(d)$ as a random draw from the distribution of $Z(d)$ conditional on $W$. For $d_1\neq d_0$, we say that $T(d_1)$ is the recanting twin of $Z(d_0)$ and viceversa. 
\end{definition}
Let
    \begin{align*}
  Y_{S_1}' &= Y(d_0, Z(d_1), M(d_1, T(d_1))),\\
  Y_{S_2}' &= Y(d_0, Z(d_0), M(d_1, T(d_1))),\\
  Y_{S_2}''& = Y(d_0, T(d_0), M(d_1, Z(d_1))),\\
  Y_{S_3}'' & = Y(d_0, T(d_0), M(d_1, Z(d_0))),
\end{align*}
define new counterfactual variables where we have replaced some counterfactual variables $Z(d)$ by their recanting twins. We can now define path-specific effects as
\begin{align*}
  \phi_{P_1} = \E(Y_{S_0} - Y_{S_1}); \quad   \phi_{P_2} = \E(Y_{S_1}' - Y_{S_2}');\quad
  \phi_{P_3} = \E(Y_{S_2}'' - Y_{S_3}'');\quad
  \phi_{P_4} = \E(Y_{S_3}  - Y_{S_4}).
  \end{align*}
  We thus have $\phi = \phi_{P_1} + \phi_{P_2} + \phi_{P_3} + \phi_{P_4} + \phi_{P_2\vee P_3}$, where $\phi_{P_2\vee P_3}$ is defined as 
\[ \phi_{P_2\vee P_3} = E(Y_{S_1} - Y_{S_1}' + Y_{S_2}' - Y_{S_2}'' +
  Y_{S_3}'' - Y_{S_3})\] 
and it measures the extent of intermediate confounding by $Z$ in the sense that $\phi_{P_2\vee P_3}=0$ whenever $Z$ is not an intermediate confounder \citep{diaz2024non}. We have the following identification result. Under model $\mathcal{M}$, let $Y(a,m,z) = f_Y(W, a, z, m, U_Y)$, $M(a,z) = f_M(W, a, z, U_M)$, and $Z(a)=f_Z(W, a, U_Z)$ denote counterfactual variables. We will assume the distribution of the errors $\P_U$ is such that the following assumptions hold.
\begin{assumption}[Sequential ignorability]\label{ass:si1} For all $(a,m,z)$:
\begin{enumerate}[label=(\roman*)]
    \item $Y(a,m,z)\indep A\mid W$; $Y(a,m,z)\indep Z\mid (A=a, W)$; and $Y(a,m,z)\indep M\mid (A=a, Z=z, W)$ 
    \item $M(a,z)\indep A\mid W$; and $M(a,z)\indep Z\mid (A=a, W)$
    \item $(Z(a),M(a))\indep A\mid W$ 
    \end{enumerate}
\end{assumption}

\begin{assumption}[Cross-world counterfactual independence]\label{ass:cw} For all $a,a', a''\in \mathrm{supp}(A)$; $z,z'\in \mathrm{supp}(Z)$ and $m\in \mathrm{supp}(M)$:
  \begin{enumerate}[label=(\roman*)]
  \item $Y(a,m,z) \indep (M(a',z'),Z(a''))\mid W;$\label{ass:cw2}
        \item $M(a,z) \indep Z(a')\mid W$.\label{ass:cw1}
        \end{enumerate}
\end{assumption}

A sufficient condition for \ref{ass:si1} and \ref{ass:cw} to hold is that all common causes of any pair of variables among $(A,M,Z,Y)$ are measured and are given by the variables that precede the earliest variable in the causal ordering of the pair. This is formalized in $\mathcal M$ as follows:
\begin{assumption}[No unmeasured confounders]\label{ass:nounc}
Assume: 
\begin{enumerate}[label=(\roman*)]
    \item $U_A\indep (U_Y, U_M, U_Z)\mid W$,
    \item $U_Z\indep (U_Y, U_M)\mid (A,W)$, and
    \item $U_M\indep U_Y\mid (Z,A,W)$.
    \end{enumerate}
\end{assumption}

\begin{theorem}[Identification of recanting twin effects under no-separability of effects]\label{theo:idenmtp}
Under model $\mathcal M$ and assumptions \ref{ass:si1} and \ref{ass:cw}, we have:

{\footnotesize
\begin{equation}\label{eq:iden}
  \begin{aligned}
    \E(Y_{S_0}\mid A=a,W=w) &=\E(Y\mid A=d_1(a,w), W=w)\notag\\
        \E(Y_{S_1}\mid a, w) &=\int\E(Y\mid d_0(a,w), z, m, w)\dd \P(m,z\mid d_1(a,w), w)\notag\\
    \E(Y_{S_1}'\mid a,w) &= \int \E(Y\mid d_0(a,w),z,m,w)\dd \P (m\mid d_1(a,w), w) \dd \P(z\mid d_1(a,w), w) \notag\\
    \E(Y_{S_2}'\mid a,w)  &= \int \E(Y\mid d_1(a,w),z,m,w)\dd \P (m\mid d_1(a,w), w) \dd \P(z\mid d_0(a,w), w) \\
    \E(Y_{S_2}''\mid a,w)  &= \int \E(Y\mid d_1(a,w),z,m,w)\dd \P (m\mid d_1(a,w), w) \dd \P(z\mid d_0(a,w), w) \\
    \E(Y_{S_3}''\mid a,w)  &= \int \E(Y\mid d_0(a,w),z,m,W) \dd \P (m\mid d_1(a,w), z',W) \dd \P(z\mid d_0(a,w), w) \dd \P(z'\mid d_0(a,w), W)\notag\\
    \E(Y_{S_3}\mid a,w)  &= \int \E(Y\mid d_0(a,w),z,m,w)\dd \P (m\mid d_1(a,w), z,w) \dd \P(z\mid d_0(a,w), w) \notag\\
        \E(Y_{S_4}\mid a,w) &=\E(Y\mid d_0(a,w), w)\notag,
\end{aligned}  
\end{equation}}%
from which we can construct identification results for $\psi_{P_j}$ and for $\psi_{P_2\vee P_3}$.     
\end{theorem}
\section{Auxiliary results}

\begin{lemma}\label{lm:natural}
    For $\psi^N$ and a fixed value $(a_1, a_2, a_3)$, define
    {\small
\begin{equation} \label{nonprimetheta_lemma1}
\begin{aligned}
\theta_3^N(a,z,m,w)&=\E(Y\mid A=a,Z=z,M=m,W=w), &     b_3^N(z,m,w;\theta_3)&=\theta_3^N(a_1,z,m,w),\\
    \theta_2^N(a,z,w)&=\E[b_3^N(Z,M,W;\theta_3^N)\mid A=a, Z=z, W=w],    & b_2^N(z,w;\theta_2)&=\theta_2^N(a_2,z,w),\\
        \theta_1^N(a,w) &= \E[b_2^N(Z,W;\theta_2^N)\mid A=a,W=w],     & b_1^N(w;\theta_1)&=\theta_1^N(a_3,w),\\
\end{aligned}
\end{equation}}
    
    as well as {\small
\begin{equation}
\begin{aligned} \label{nonprimealpha_lemma1}
         \alpha_1^N(a,w) & = \frac{\one(A=a_3)}{\P(a\mid w)},\\
        \alpha_2^N(a,z,w) & = \frac{\one(A=a_2)}{\P(a\mid w)}\frac{\P(z\mid A=a_3, w)}{\P(z\mid a, w)},\\
        \alpha_3^N(a,z,m,w) & = \frac{\one(A=a_1)}{\P(a\mid w)}\frac{\P(m\mid A=a_2, z, w)}{\P(m\mid a, z, w)}\frac{\P(z\mid A=a_3, w)}{\P(z\mid a, w)}.
\end{aligned}
\end{equation}}%
{\small
\begin{equation}
\begin{aligned}
\Psi_0^N(\theta_0^N) & = \theta_0^N\\
\Psi_1^N(\theta_1^N) &= \E\{b_{1}^N(W;\theta_{1}^N)\};\\
\Psi_2^N(\theta_2^N) &= \E[\E\{b_{2}^N(Z,W;\theta_{2}^N)\mid A=a_3, W\}];\\
\Psi_3^N(\theta_3^N) &= \E\big[\E\big\{\E[b_{3}^N(Z,M,W;\theta_{3}^N)\mid A=a_2,Z, W]\mid A=a_3, W\big\}\big].
\end{aligned}\label{eq:theta_lemma1}
\end{equation}
}

For $\psi^R$ and a fixed value $(a_1,a_2,a_3,a_4)$, define
{\small
\begin{equation} \label{primetheta-proof}
\begin{aligned}
\theta^R_4(a,z,m,w)&=\E(Y\mid A=a,Z=z,M=m,W=w), &b^R_4(z,m,w;\theta_4^R)&=\theta^R_4(a_1,z,m,w),\\
\theta^R_{3}(a,m,w)&=\E[b^R_4(\Zp,M,W;\theta_4^R)\mid A=a, M=m, W=w], &b^R_{3}(m,w;\theta_{3}^R)&=\theta^R_{3}(a_2,m,w), \\
\theta^R_{2}(a,z,w)&=\E[b^R_{3}(M,W;\theta_{3}^R)\mid A=a, Z=z, W=w], & b^R_{2}(z,w;\theta_{2}^R)&=\theta^R_{2}(a_3,z,w), \\
\theta^R_{1}(a,w) &= \E[b^R_{2}(Z,W;\theta_{2}^R)\mid A=a,W=w], & b^R_{1}(w;\theta_{1}^R)&=\theta^R_{1}(a_4,w),  \\
\end{aligned}
\end{equation}}%
as well as
{\small
\begin{equation} 
\begin{aligned}
\alpha^R_{1}(a,w) & = \frac{\one(A=a_4)}{\P(a\mid w)};\\
\alpha^R_{2}(a,z,w) & = \frac{\one(A=a_3)}{\P(a\mid w)}\frac{\P(z\mid A=a_4, w)}{\P(z\mid a, w)};\\
\alpha^R_{3}(a,m,w) & = \frac{\one(A=a_2)}{\P(a\mid w)}\frac{\int_z \P(m\mid A=a_3, z, w) \dd \P(z\mid A=a_4,w) }{\P(m\mid a, w)};\\
\alpha^R_4(a,z,m,w) & = \frac{\one(A=a_1)}{\P(a\mid w)}\frac{\P(z\mid A=a_2, w)}{\P(z\mid a, w)}\frac{\int_z \P(m\mid A=a_3, z, w) \dd \P(z\mid A=a_4,w) }{\P(m\mid a, z, w)},
\end{aligned}
\end{equation}}%
{\small
\begin{equation}
\begin{aligned}
\Psi_0^R(\theta_0^R) & = \theta_0^R \\
\Psi_1^R(\theta_1^R) & = \E [b_{1}^R(W;\theta_{1}^R) ] \\
 \Psi_2^R(\theta_2^R) &= \E[\E\{b_{2}^R(Z,W;\theta_{2}^R)\mid A=a_4, W\}] \\
 \Psi_3^R(\theta_3^R) &= \E[\E\{\E[b_{3}^R(M,W;\theta_{3}^R)\mid A=a_3,Z, W]\mid A=a_4, W\}] \\
 \Psi_4^R(\theta_4^R) &= \E \{ \E[\E\{\E [b_4^R(Z^\pi, M,W;\theta_4^R)\mid A=a_2,M, W]\mid A=a_3, Z, W\}\mid A=a_4, W] \},
\end{aligned}\label{eq:prime_estimator}
\end{equation}
}
Then \[\Psi_j^E(\theta_j^E)=\E[\alpha_{j}^E(X)\theta_j^E(X)]\] for all $E \in \{N, R \}$ and $\theta_j$.
\end{lemma}

\textit{Proof} \quad We will present the proof for $\psi^N$ and $\psi^R$ separately. Without affecting the clarity, we ignore the superscript $N$ or $R$ in the following proof for simplicity.

For $\psi^N$, the proof is done by noting that 

\begin{align*}
\E \left[\alpha_1 (X) \theta_1 (X)\right] & = \E \left[\alpha_1 (A, W) \theta_1 (A, W)\right]  = \E \left[ \frac{\one (A = a_3)}{\P(A = a_3 \mid W)} \theta_1 (A, W) \right] \\ & = \E \left[ \frac{\one (A = a_3)}{\P(A = a_3 \mid W)} b_1 (W; \theta_1) \right] 
 = \E \left\{ \E \left[ \frac{\one (A = a_3)}{\P(A = a_3 \mid W)} b_1 (W; \theta_1) \right] \bigg| ~ W \right\} \\ & = \E \left[ b_1 (W; \theta_1) \right] = \Psi_1(\theta_1),
\end{align*}

\begin{align*}
     \E[\alpha_2 (X) \theta_2 (X)] & = \E[\alpha_2 (A,Z, W) \theta_2 (A,Z, W)]  = \E \left[ \frac{\one (A = a_2) }{\P(A = a_2 \mid W)} \frac{\P(Z \mid A = a_3, W)}{\P(Z \mid A, W)} \theta_2 (A,Z, W) \right] \\ 
    & = \E \left[ \frac{\one (A = a_2) }{\P(A = a_2 \mid W)} \frac{\P(Z \mid A = a_3, W)}{\P(Z \mid A, W)} b_2 (Z, W; \theta_2) \right] \\
    & = \E \left\{ \E \left[ \frac{\one (A = a_2) }{\P(A = a_2 \mid W)} \frac{\P(Z \mid A = a_3, W)}{\P(Z \mid A, W)} b_2 (Z, W; \theta_2) ~ \bigg| ~ A, W \right] \right\} \\
    & = \E \left[  \frac{\one (A = a_2) }{\P(A = a_2 \mid W)} \P(Z \mid A = a_3, W) b_2 (Z, W; \theta_2) \right] \\
    & = \E \left\{ \E \left[  \frac{\one (A = a_2) }{\P(A = a_2 \mid W)} \P(Z \mid A = a_3, W) b_2 (Z, W; \theta_2) 
 ~ \bigg| ~ W \right] \right\} \\
     & =  \E [ \one (A = a_2) \P(Z \mid A = a_3, W) b_2 (Z, W; \theta_2)  ]  \\
    & = \E \{ \E[b_2(Z, W; \theta_2) \mid A = a_3, W]\} = \Psi_2(\theta_2), \\ 
\end{align*}
and
\begin{align*}
    & \E \left[\alpha_3 (X) \theta_3 (X) \right] = \E \left[\alpha_3 (A,Z, M, W) \theta_3 (A,Z, M, W) \right]  \\ & = \E \left[ \frac{\one (A = a_1) }{\P(A = a_1 \mid W)} \frac{\P(M \mid A = a_2, Z, W)}{\P(M \mid A, Z, W)} \frac{\P(Z \mid A = a_3, W)}{\P(Z \mid A, W)} \theta_3 (A,Z, M, W) \right] \\ 
    & = \E \left[ \frac{\one (A = a_1) }{\P(A = a_1 \mid W)} \frac{\P(M \mid A = a_2, Z, W)}{\P(M \mid A, Z, W)} \frac{\P(Z \mid A = a_3, W)}{\P(Z \mid A, W)} b_3 (Z, M, W; \theta_3) \right] \\
    & = \E \left\{ \E \left[ \frac{\one (A = a_1) }{\P(A = a_1 \mid W)} \frac{\P(M \mid A = a_2, Z, W)}{\P(M \mid A, Z, W)} \frac{\P(Z \mid A = a_3, W)}{\P(Z \mid A, W)} b_3 (Z, M, W; \theta_3) ~\bigg| ~ A, Z, W \right] \right\} \\
    & = \E \left[  \frac{\one (A = a_1) }{\P(A = a_1 \mid W)} {\P(M \mid A = a_2, Z, W)} \frac{\P(Z \mid A = a_3, W)}{\P(Z \mid A, W)} b_3 (Z, M, W; \theta_3)  \right] \\
    & = \E[  {\one (A = a_1) } {\P(M \mid A = a_2, Z, W)} {\P(Z \mid A = a_3, W)} b_3 (Z, M, W; \theta_3)   ] \\
    & = \E[\E\{\E [b_3(Z,M,W;\theta_3)\mid A=a_2,Z, W]\mid A=a_3, W\}] = \Psi_3(\theta_3).
\end{align*}

For $\psi^R$, the proof is done by noting that 

\begin{align*}
\E[\alpha_1 (X) \theta_1 (X)] & = \E[\alpha_1 (A, W) \theta_1 (A, W)] = \E \left[ \frac{\one (A = a_4)}{\P(A = a_4 \mid W)} \theta_1 (A, W) \right] \\ & = \E \left[ \frac{\one (A = a_4)}{\P(A = a_4 \mid W)} b_1 (W; \theta_1) \right] 
 = \E \left\{ \E \left[ \frac{\one (A = a_4)}{\P(A = a_4 \mid W)} b_1 (W; \theta_1) \right] ~ \bigg| ~ W \right\} \\ & = \E[ b_1 (W; \theta_1)] = \Psi_1(\theta_1),
\end{align*}

\begin{align*}
   \E[\alpha_2 (X) \theta_2 (X)]  & = \E[\alpha_2 (A,Z, W) \theta_2 (A,Z, W)]  \\ & = \E \left[ \frac{\one (A = a_3) }{\P(A = a_3 \mid W)} \frac{\P(Z \mid A = a_4, W)}{\P(Z \mid A, W)} \theta_2 (A,Z, W) \right] \\ 
    & = \E \left[ \frac{\one (A = a_3) }{\P(A = a_3 \mid W)} \frac{\P(Z \mid A = a_4, W)}{\P(Z \mid A, W)} b_2 (Z, W; \theta_2) \right] \\
    & = \E \left\{ \E \left[ \frac{\one (A = a_3) }{\P(A = a_3 \mid W)} \frac{\P(Z \mid A = a_4, W)}{\P(Z \mid A, W)} b_2 (Z, W; \theta_2) ~ \bigg| ~ A, W \right] \right\} \\
    & = \E \left[  \frac{\one (A = a_3) }{\P(A = a_3 \mid W)} \P(Z \mid A = a_4, W) b_2 (Z, W; \theta_2) \right] \\
    & = \E \left\{ \E \left[  \frac{\one (A = a_3) }{\P(A = a_3 \mid W)} \P(Z \mid A = a_4, W) b_2 (Z, W; \theta_2) 
 ~ \bigg| ~ W \right] \right\} \\
     & =  \E [ \one (A = a_3) \P(Z \mid A = a_4, W) b_2 (Z, W; \theta_2)  ]  \\
    & = \E \{\E [b_2(Z, W; \theta_2) \mid A = a_4, W]\} = \Psi_2(\theta_2),
\end{align*}

\begin{align*}
    & \E[\alpha_{3}(X)\theta_{3}(X)] = \E[\alpha_{3}(A,M,W)\theta_{3}(A,M,W)] \\
    & = \E \left[ \frac{\one(A=a_2)}{\P(A\mid W)}\frac{\int \dd \P(M\mid A=a_3, Z, W)\dd \P(Z\mid A=a_4,W)}{\P(M\mid A, W)}\ \theta_{3}(A,M,W) \right] \\ 
    & = \E \left[\frac{\one(A=a_2)}{\P(A\mid W)}\frac{\int \dd \P(M\mid A=a_3, Z, W) \dd \P(Z\mid A=a_4,W)}{\P(M\mid A, Z, W)\P(Z \mid A, W)}\ \theta_{3}(A,M,W) \right]  \\ 
    & = \E \left[\frac{\one(A=a_2)}{\P(A\mid W)}\frac{\int \dd \P(M\mid A=a_3, Z, W) \dd \P(Z\mid A=a_4,W)}{\P(M\mid A, Z, W)\P(Z \mid A, W)}\ b_{3}(M,W; \theta_3) \right]  \\ 
    & = \E \left\{ \E \left[ \frac{\one(A=a_2)}{\P(A\mid W)}\frac{\int  \dd \P(M\mid A=a_3, Z, W) \dd \P(Z\mid A=a_4,W)}{\P(M\mid A, Z, W)\P(Z \mid A, W)}\ b_{3}(M,W; \theta_3)~ \bigg| ~ A, Z, W \right] \right\} \\ 
    & = \E \left[ \frac{\one(A=a_2)}{\P(A\mid W)}\frac{\int \dd \P(M\mid A=a_3, Z, W) \dd \P(Z\mid A=a_4,W)}{\P(Z \mid A, W)}\ b_{3}(M,W; \theta_3)  \right] \\ 
    & = \E \left[ {\one(A=a_2)}{b_{3}(M,W; \theta_3) \int \dd \P(M\mid A=a_3, Z, W)\dd \P(Z\mid A=a_4,W)}   \right] \\ 
    & = \E\big[\E\big\{\E[b_{3}(M,W;\theta_{3})\mid A=a_3,Z, W]\mid A=a_4, W\big\}\big] = \Psi_3(\theta_3),\\
\end{align*}

and

\begin{align*}
    & \E[\alpha_{4}(X)\theta_{4}(X)] = \E[\alpha_{4}(A,Z,M,W)\theta_{4}(A, Z, M, W)] \\
    & = \E \left[ \frac{\one(A=a_1)}{\P(A\mid W)}\frac{\P(Z\mid A=a_2, W)}{\P(Z\mid A, W)}\frac{\int \dd \P(M\mid A=a_3, Z, W)\dd \P(Z\mid A=a_4,W) }{\P(M\mid A, Z, W)}\theta_{4}(A, Z, M, W) \right] \\ 
    & = \E \left[\frac{\one(A=a_1)}{\P(A\mid W)}\frac{\P(Z\mid A=a_2, W)}{\P(Z\mid A, W)}\frac{\int \dd \P(M\mid A=a_3, Z, W) \dd \P(Z\mid A=a_4,W) }{\P(M\mid A, Z, W)}b_{4}(Z, M, W; \theta_4) \right] \\ 
    & = \E \left[ {\one(A=a_1)} b_{4}(Z, M, W; \theta_4) {\P(Z\mid A=a_2, W)}{\int \dd \P(M\mid A=a_3, Z, W) \dd \P(Z\mid A=a_4,W)} \right] \\ 
    & = \E \left[ {\one(A=a_1)} b_{4}(Z^\pi, M, W; \theta_4) {\P(Z^\pi \mid A=a_2, M, W)}{\int \dd \P(M\mid A=a_3, Z, W) \dd \P(Z\mid A=a_4,W)} \right] \\ 
    & = \E \left\{ \E\big[\E\big\{\E [b_4(Z^\pi, M,W;\theta_4)\mid A=a_2,M, W]\mid A=a_3, Z, W\big\}\mid A=a_4, W\big] \right\} = \Psi_4(\theta_4).\\
\end{align*}
$\square$

\section{Proofs of Proposition 1}
For natural effects \citep{pearl2001direct}, we define 
\[
\psi^{natural}(a_1, a_2) = \int_w \int_m \E(Y\mid A=a_1, m, w) \dd \P (m \mid A = a_2, w) \dd \P (w), \\ 
\]
then $\operatorname{NIE} = \psi^{natural}(1, 1) - \psi^{natural}(1, 0)$, and $\operatorname{NDE} = \psi^{natural}(1, 0) - \psi^{natural}(0, 0)$. Note that 
\begin{align*}
    \psi^{natural}(a_1, a_2) &= \int_w \int_m \E(Y\mid A=a_1, m, w) \dd \P (m \mid A = a_2, w) \dd \P (w) \\ 
    & = \int_w \int_m \E(Y\mid A=a_1, z, m, w) \dd \P (m \mid A = a_2, w) \dd \P (w) \quad (\text{$Z$ is null})  \\ 
    & = \int_w \int_z \int_m \E(Y\mid A=a_1, z, m, w) \dd \P (m \mid A = a_2, z, w) \dd \P (z \mid A = a_2, w) \dd \P (w) \\ 
    & = \E \bigg[\E \Big[ \E \big[ \E(Y\mid A=a_1, Z, M, W) \,\big|\, A = a_2, Z, W\big] \,\Big|\, A = a_2, W\Big]\bigg], \\ 
    & = \psi^{N}(a_1, a_2, a_2).
    \end{align*}
This shows $\operatorname{NDE} = \psi^N (1, 0 , 0) - \psi^N (0, 0 , 0)$ and $\operatorname{NIE} = \psi^N (1, 1 , 1) - \psi^N (1, 0 , 0)$. \\
We skip the decision theoretic effects \citep{geneletti2007identifying} because they share the same identification formulas with natural effects. For the organic effects \citep{lok2015organic}, note that they are identified as 

\begin{equation}
    \operatorname{ODE} = \psi^{organic}(0, 1)- \psi^{organic}(0, 0), \operatorname{OIE} = \psi^{organic}(1, 1) - \psi^{organic}(0, 1),
\end{equation}
where 
\[
\psi^{organic}(a_1, a_2)  = \int_w \int_z \int_m \E[Y \mid A = a_1, M = m, Z = z, W = w] \dd \P(m \mid a_2, z, w) \dd \P (z \mid a_1, w) \dd \P(w),
\]
the equations $\operatorname{ODE} = \psi^N (0, 1 , 0) - \psi^N (0, 0 , 0)$ and $\operatorname{OIE} = \psi^N (1, 1 , 1) - \psi^N (0, 1 , 0)$ became obvious after we noticed that $\psi^{organic}(a_1, a_2) = \psi^N(a_1, a_2, a_1)$. \\
For the randomized interventional effects \citep{vanderweele2014effect}, we define 
\[
\psi^{randomized}(a_1, a_2) =  \int_w \int_m \int_z \E (Y \mid A = a_1, z, m, w) \dd \P(z \mid A = a_1, w) \dd \P(m \mid A = a_2, w)  \dd \P (w),
\]
then $\operatorname{RIIE} = \psi^{randomized} (1, 1) - \psi^{randomized} (1, 0)$ and $\operatorname{RIDE} = \psi^{randomized} (1, 0) - \psi^{randomized} (0, 0)$. Note that
{\footnotesize
\begin{align*}
    \psi^{randomized}(a_1, a_2) & =  \int_w \int_m \int_z \E (Y \mid A = a_1, z, m, w) \dd \P(z \mid A = a_1, w) \dd \P(m \mid A = a_2, w)  \dd \P (w) \\
    & =  \int_w \int_z \int_m \int_{z'} \E(Y\mid a_1,Z = z',m,w) \dd \P(z' \mid a_1, w) \dd \P (m\mid a_2, Z = z,w) \dd \P(z\mid a_2, w)\dd\P(w) \\
    & =  \int_w \int_z \int_m \int_{z'} \E(Y\mid a_1,Z^\pi = z',m,w) \dd \P(z' \mid a_1, w) \dd \P (m\mid a_2, Z = z,w) \dd \P(z\mid a_2, w)\dd\P(w) \\
    & =  \int_w \int_z \int_m \int_{z'} \E(Y\mid a_1,Z^\pi = z',m,w) \dd \P(z' \mid a_1, m, w) \dd \P (m\mid a_2, Z = z,w) \dd \P(z\mid a_2, w)\dd\P(w) \\
    & = \E\Bigg[\E\bigg[\E\Big[\E \big[ \E(Y\mid a_1,Z^\pi ,M,W) \,\big|\, A=a_1,M, W\big ]\,\Big|\, A=a_2, Z, W\Big]\,\bigg|\, A=a_2, W\bigg]\Bigg],\\
    & = \psi^R(a_1, a_1, a_2, a_2),
\end{align*}
}
where the third and fourth equalities hold because $Z^\pi \mid A, W \sim Z \mid A, W$ and $Z^\pi \indep M \mid A, W$, respectively. This shows $\operatorname{RIDE} = \psi^R(1, 1, 0, 0) - \psi^R(0, 0, 0, 0)$ and $\operatorname{RIIE} = \psi^R(1, 1, 1, 1) - \psi^R(1, 1, 0, 0)$. 

For the path-specific effects based on recanting twins \citep{diaz2024non}, we note that $\operatorname{RT}_1 = E(Y_{S_0} - Y_{S_1})$, where 
\begin{align*}
    E(Y_{S_0}) & = \int_w E(Y \mid A = 1, w) \dd \P(w) \\
    & = \int_w \int_z \int_m E(Y \mid A = 1, z, m, w) \dd \P (m \mid A= 1, z, w) \dd \P (z \mid A = 1, w)  \dd \P(w) \\
    & = \psi^N(1, 1,1),
\end{align*}
and
\begin{align*}
    E(Y_{S_1}) & = \int_w \int_z \int_m E(Y \mid A = 0, z, m, w) \dd \P (z, m \mid A= 1, w) \dd \P(w) \\
    & = \int_w \int_z \int_m E(Y \mid A = 0, z, m, w) \dd \P (m \mid A= 1, z, w) \dd \P (z \mid A = 1, w)  \dd \P(w) \\
    & = \psi^N(0, 1,1).
\end{align*}
For $\operatorname{RT}_2$, we note that $\operatorname{RT}_2 = E(Y_{S_1}' - Y_{S_2}')$, where 
{\footnotesize
\begin{align*}
    E(Y_{S_1}') & = \int_w \int_z \int_m E(Y \mid A = 0, z, m, w) \dd \P (m \mid A= 1, w) \dd \P (z \mid A= 1, w) \dd \P(w) \\
    & =  \int_w \int_z \int_m \int_{z'} \E(Y\mid 0,Z^\pi = z',m,w) \dd \P(z' \mid 1, m, w) \dd \P (m\mid 1, Z = z,w) \dd \P(z\mid 1, w)\dd\P(w) \\
    & = \psi^R(0, 1, 1,1),
\end{align*}
}
and
{\footnotesize
\begin{align*}
    E(Y_{S_2}') & = \int_w \int_z \int_m E(Y \mid A = 0, z, m, w) \dd \P (m \mid A= 1, w) \dd \P (z \mid A= 0, w) \dd \P(w) \\
    & =  \int_w \int_z \int_m \int_{z'} \E(Y\mid 0,Z^\pi = z',m,w) \dd \P(z' \mid 0, m, w) \dd \P (m\mid 1, Z = z,w) \dd \P(z\mid 1, w)\dd\P(w) \\
    & = \psi^R(0, 0, 1,1).
\end{align*}
}
For $\operatorname{RT}_3$, we note that $\operatorname{RT}_3 = E(Y_{S_2}'' - Y_{S_3}'')$, where 
\[
    E(Y_{S_2}'') = E(Y_{S_2}') = \psi^R(0, 0, 1,1),
\]
and
{\footnotesize
\begin{align*}
    E(Y_{S_3}'') & = \int_w \int_z \int_m \int_{z'} E(Y \mid A = 0, z, m, w) \dd \P (m \mid A= 1, z', w) \dd \P (z \mid A= 0, w)  \P (z' \mid A= 0, w) \dd \P(w) \\
    & =  \int_w \int_z \int_m \int_{z'} \E(Y\mid 0,Z^\pi = z',m,w) \dd \P(z' \mid 0, m, w) \dd \P (m\mid 1, Z = z,w) \dd \P(z\mid 0, w)\dd\P(w) \\
    & = \psi^R(0, 0, 1,0).
\end{align*}
}
At last, for $\operatorname{RT}_4$, we note that $\operatorname{RT}_4 = E(Y_{S_3} - Y_{S_4})$, where 
\begin{align*}
    E(Y_{S_3}) & = \int_w\int_z \int_m E(Y \mid A = 0, z, m, w) \dd \P (m \mid A = 1, z, w) \dd \P (z \mid A = 0, w) \dd \P(w) \\
    & = \psi^N(0, 1,0),
\end{align*}
and 
\begin{align*}
    E(Y_{S_4}) & = \int_w E(Y \mid A = 0, w)  \dd \P(w) \\
    & = \int_w \int_z \int_m E(Y \mid A = 0, z, m, w) \dd \P (m \mid A= 0, z, w) \dd \P (z \mid A = 0, w)  \dd \P(w) \\
    & = \psi^N(0, 0, 0).
\end{align*}
Therefore, we show that 
{\footnotesize
\begin{align*}
    \operatorname{RT}_1 &= \psi^N(1,1,1)-\psi^N(0,1,1), &\operatorname{RT}_2 &= \psi^R(0,1,1,1) - \psi^R(0,0,1,1),\\
    \operatorname{RT}_3 &= \psi^R(0,0,1,1) - \psi^R(0,0,1,0), &\operatorname{RT}_4 &= \psi^N(0,1,0) - \psi^N(0,0,0),
\end{align*}}%

We skip the path-specific effects based on separable effects \citep{vo2024recanting} as they share the same identification formulas with the path-specific effects based on recanting twins. 
\section{Proofs of Theorem 1}

We will present the proof for $\psi^N$. The proof for $\psi^R$ follows symmetrical arguments and is therefore omitted. We omit the index $N$ in the following notation for simplicity. 

For two arbitrary distributions $\F$ and $\P$, define
\begin{align*}
\theta_{3,\P}(a,z,m,w) &= \E_\P(Y\mid A=a, Z=z, M=m, W=w),\\
    \theta_{2,\P,\F}(a,z,w)&=\E_\P[b_3(Z,M,W;\theta_{3,\F})\mid A=a, Z=z, W=w],\\
        \theta_{1,\P,\F}(a,w) &= \E_\P[b_2(Z,W;\theta_{2,\F})\mid A=a,W=w],\\
        \theta_{0,\P,\F}&= \E_\P[b_1(W;\theta_{1,\F})],
\end{align*}
where for $\P=\F$ we simply denote $\theta_{j,\P}= \theta_{j,\P,\P}$. Define the linear functionals
\begin{equation}
\begin{aligned}
\Psi_0(\theta_0) & = \theta_0\\
\Psi_1(\theta_1) &= \E_\P\{b_{1}(W;\theta_{1})\};\\
\Psi_2(\theta_2) &= \E_\P[\E_\P\{b_{2}(Z,W;\theta_{2})\mid A=a_3, W\}];\\
\Psi_3(\theta_3) &= \E_\P\big[\E_\P\big\{\E_\P[b_{3}(Z,M,W;\theta_{3})\mid A=a_2,Z, W]\mid A=a_3, W\big\}\big].
\end{aligned}\label{eq:theta}
\end{equation}
First, notice that for $j=1,2,3$, $\Psi_j(\theta_{j,\F}) = \Psi_{j-1}(\theta_{j-1,\P,\F})$. This yields
\[\Psi_{0}(\theta_{0,\F}) - \Psi_{3}(\theta_{3,\P})  = - \sum_{j=0}^3\{\Psi_{j}(\theta_{j,\P,\F}) - \Psi_{j}(\theta_{j,\F})\}.\]
According to the Riesz representation theorem, there exists a function $\alpha_{j,\P}$ such that 
\[\Psi_j(\theta_j)=\E_{\P}[\alpha_{j,\P}(X)\theta_j(X)]\]
for all $\theta_j$. According to Lemma~\ref{lm:natural}, such $\alpha_{j,\P}$ functions are given by the formulas in (\ref{nonprimealpha}) and (\ref{primealpha}) of the main text and $\alpha_{0,\P}=1$. Thus, 
\begin{align*}
    \Psi_{j}(\theta_{j,\P,\F}) - \Psi_{j}(\theta_{j,\F}) & = \E_\P[\alpha_{j,\P}(X)\{\theta_{j,\P,\F}(X) - \theta_{j,\F}(X)\}]\\
    & = \E_\P[\{\alpha_{j,\P}(X)-\alpha_{j,\F}(X)\}\{\theta_{j,\P,\F}(X) - \theta_{j,\F}(X)\}]\\
    &+\E_\P[\alpha_{j,\F}(X)\{b_{j+1}(X;\theta_{j+1, \F}) - \theta_{j,\F}(X)\}],
\end{align*}
where we define $b_4(X; \theta_4) = Y$ and the second equality follows by adding and subtracting $\alpha_{j,\F}$ and by iterated expectation. Define 
\[R_2(\eta_\F,\eta_\P) = -\sum_{j=0}^3\E_\P[\{\alpha_{j,\P}(X)-\alpha_{j,\F}(X)\}\{\theta_{j,\P,\F}(X) - \theta_{j,\F}(X)\}],\]
and
\[
\varphi(X; \eta_\F) = \sum_{j = 0}^3 \alpha_{j,\F}(X)\{b_{j+1}(X;\theta_{j+1, \F}) - \theta_{j,\F}(X)\}
\]
Then $\varphi^N(X; \eta^N)$ is the gradient of the first-order von Mises expansion for $\psi^N$ because $\Psi_0(\theta_{0,\F})=\psi(\F)$ and $\Psi_3(\theta_{3,\P})=\psi(\P)$.

Next, we prove that $\var[\varphi^{N}(X; \P)]$ is the corresponding non-parametric efficiency bound in a model with observed data $X=(W, A, Z, \Zp, M, Y)$, this holds if 
\begin{equation} \label{eq:sx}
    \frac {\dd}{\dd \epsilon} \psi(\P_\epsilon) \bigg|_{\epsilon = 0} = \E [\varphi(X) s(X)],
\end{equation}
where $\P_\epsilon$ is a smooth parametric submodel that locally covers the non-parametric model, and the score $s(X)$ is defined as 
\[
s(X) = \left( \frac{\dd \log \P_\epsilon}{\dd \epsilon} \right)\bigg|_{\epsilon = 0},
\]
where $\P_{\epsilon = 0} = \P$. 

To validate (\ref{eq:sx}), we compute the bias term $\psi(\P_\epsilon) - \psi(\P)$.
Note that 
\begin{align} \label{eq:differentiate}
\psi(\P_\epsilon) - \psi(\P) & = \Psi_{0}( \theta_{0, \P_{\epsilon}}) - \Psi_{3}(\theta_{3, \P}) \notag \\
    & = R_2(\eta_{\P_\epsilon}, \eta_\P) - \E_\P [\varphi(X; \eta_{\P_\epsilon})] \notag \\
    & = \E_{\P_{\epsilon}} [\varphi(X; \eta_\P)] - R_2(\eta_{\P}, \eta_{\P_\epsilon}),
\end{align}

% where we define $b_4(X; \hat \theta_4) = Y$. The first equality is obtained by noticing $\Psi_j (\hat \theta_j) = \Psi_{j - 1} (\tilde \theta_{j - 1}), j = 2, 3$; the second equality is obtained by the facts that $\Psi_{j}(\tilde \theta_{j}) - \Psi_{j}(\hat\theta_{j}) = \E[\alpha_{j}(X)\{\tilde \theta_{j}(X) - \hat\theta_{j}(X)\}]$ and $\E [\hat \alpha_j(X) \tilde \theta_j(X)] = \E [\hat \alpha_j(X) b_{j+1}(X; \hat \theta_{j + 1})]$ for $j = 1, 2, 3$, and the last equality is obtained by Theorem \ref{theo:vonm} in the main text and exchanging the role of $\P$ and $\P_\epsilon$. 

Differentating with respect to $\epsilon$ on equation (\ref{eq:differentiate}) and evaluating at $\epsilon = 0$ yields

\begin{align*}
    \frac{\dd}{\dd \epsilon} \psi(\P_\epsilon) & = \int \varphi(X; \eta_\P) \left( \frac {\dd \P_\epsilon}{\dd \epsilon} \right) \bigg|_{\epsilon = 0} - \frac{\dd}{\dd \epsilon} R_2 (\eta_\P, \eta_{\P_\epsilon}) \bigg|_{\epsilon = 0} \\
    & = \int \underbrace{\varphi(X; \eta_\P)}_{\varphi(X)} \underbrace{\left( \frac {\dd \log \P_\epsilon}{\dd \epsilon} \right) \bigg|_{\epsilon = 0}}_{s(X)} \dd \P- \frac{\dd}{\dd \epsilon} R_2 (\eta_\P, \eta_{\P_\epsilon}) \bigg|_{\epsilon = 0}.
\end{align*}

The proof is done (equation (\ref{eq:sx}) is satisfied) by noting that $\frac{\dd}{\dd \epsilon} R_2 (\eta_\P, \eta_{\P_\epsilon}) |_{\epsilon = 0} = 0$ as this is a second order term. $\square$

% \section{Proofs of Proposition 1}\label{sec:delete}

% \id{I think we can delete this proof with the current paper setup.}
% Note that \begin{align*}
%     \psi_{a_1a_2}^N &= \int_w \int_m \E(Y\mid A=a_1, m, w) \dd \P (m \mid A = a_2, w) \dd \P (w) \\ 
%     & = \int_w \int_m \E(Y\mid A=a_1, z, m, w) \dd \P (m \mid A = a_2, w) \dd \P (w) \quad (\text{$Z$ is null})  \\ 
%     & = \int_w \int_z \int_m \E(Y\mid A=a_1, z, m, w) \dd \P (m \mid A = a_2, z, w) \dd \P (z \mid A = a_2, w) \dd \P (w) \\ 
%     & = \psi_{a_1a_2a_2}^{N}
%     \end{align*}
% \begin{align*}
%     \psi^I_{a_1a_2} & =  \int_w \int_m \int_z \E (Y \mid A = a_1, z, m, w) \dd \P(z \mid A = a_1, w) \dd \P(m \mid A = a_2, w)  \dd \P (w) \\
%     & =  \int_w \int_z \int_m \int_{z'} \E(Y\mid a_1,Z = z',m,w) \dd \P(z' \mid a_1, w) \dd \P (m\mid a_2, Z = z,w) \dd \P(z\mid a_2, w)\dd\P(w) \\
%     & = \psi^R_{a_1a_1a_2a_2}.
% \end{align*}

% Besides, $\psi^O_{a_1,a_2} = \psi_{a_1a_2a_1}^{N}$, $\psi_{a_1a_2a_3}^N = \psi^{N}_{a_1a_2a_3}$, and $\psi_{a_1a_2a_3a_4}^R = \psi^R_{a_1a_2a_3a_4}$ are trivial. This ends the proof.
% $\square$

% \id{Please change the below title for Proof of Proposition~\ref{prop:loss_alpha}}
\section{Proofs of Proposition 2}

% \subsection{natural parameter $\psi^N$} \label{sec:nonprime}
We will present the proof for $\psi^N$. The proof for $\psi^R$ follows symmetrical arguments and is therefore omitted. We omit the index $N$ in the following notation for simplicity.

Note that 
\begin{align*}
\psi^N &= \E\{b_1(W;\theta_1)\};\\
 &= \E[\E\{b_2(Z,W;\theta_2)\mid A=a_3, W\}];\\
 &= \E\big[\E\big\{\E [b_3(Z,M,W;\theta_3)\mid A=a_2,Z, W]\mid A=a_3, W\big\}\big],\label{eq:nonprime}
\end{align*}
where we defined
\begin{align*}
\theta_3(a,z,m,w)&=\E(Y\mid A=a,Z=z,M=m,W=w);\\
    \theta_2(a,z,w)&=\E[b_3(Z,M,W;\theta_3)\mid A=a, Z=z, W=w];\\
        \theta_1(a,w) &= \E[b_2(Z,W;\theta_2)\mid A=a,W=w],
\end{align*}
\begin{equation}
\begin{aligned}
        \alpha_1(A,W) & = \frac{\one(A=a_3)}{\P(A\mid W)};\\
        \alpha_2(A,Z,W) & = \frac{\one(A=a_2)}{\P(A\mid W)}\frac{\P(Z\mid A=a_3, W)}{\P(Z\mid A, W)};\\
        \alpha_3(A,Z,M,W) & = \frac{\one(A=a_1)}{\P(A\mid W)}\frac{\P(M\mid A=a_2, Z, W)}{\P(M\mid A, Z, W)}\frac{\P(Z\mid A=a_3, W)}{\P(Z\mid A, W)},.
\end{aligned}\label{eq:natural_alpha123}    
\end{equation}
and
\begin{align*}
    b_3(z,m,w;\theta_3)&=\theta_3(a_1,z,m,w);\\
    b_2(z,w;\theta_2)&=\theta_2(a_2,z,w);\\
    b_1(w;\theta_1)&=\theta_1(a_3,w).
\end{align*}
By Lemma \ref{lm:natural}, 
\begin{equation}
\begin{aligned}
 \E\{b_1(W;\theta_1)\} & = \E[\alpha_1(A,W)\theta_1(A,W)];\\
 \E[\E\{b_2(Z,W;\theta_2)\mid A=a_3, W\}] & = \E[\alpha_2(A,Z,W)\theta_2(A,Z,W)];\\
 \E[\E\{\E [b_3(Z,M,W;\theta_3)\mid A=a_2,Z, W]\mid A=a_3, W\}] & = \E[\alpha_3(A,Z,M,W)\theta_3(A,Z,M,W)].
\end{aligned}\label{eq:natural_b1b2b3}
\end{equation}

In the following we do not assume we know the functional forms of $(\alpha_1, \alpha_2, \alpha_3)$, and instead propose loss functions to estimate them directly.
Starting from the mean-squared error (MSE) loss function, we have 

\begin{align}
    \alpha_1 &= \arg\min_{\tilde\alpha_1}\E[(\alpha_1(A, W) - \tilde\alpha_1(A, W))^2]\notag\\
    &= \arg\min_{\tilde\alpha_1}\E[\alpha_1(A, W)^2 - 2\tilde\alpha_1(A, W)\alpha_1(A, W)+\tilde\alpha_1(A, W)^2]\notag\\
        &= \arg\min_{\tilde\alpha_1}\E[\tilde\alpha_1(A, W)^2 - 2\tilde\alpha_1(A, W)\alpha_1(A, W)]\notag\\
                                &= \arg\min_{\tilde\alpha_1}\E[\tilde\alpha_1(A, W)^2 - 2b_1(W, \tilde\alpha_1)],\label{eq:alpha1}
\end{align}
where the last equality follows from the first equation in (\ref{eq:natural_b1b2b3}). Likewise, for $\alpha_2$, we can get 
\begin{align}
    \alpha_2  & =  \arg\min_{\tilde\alpha_2}\E\{\tilde\alpha_2(A, Z, W)^2 - 2\E[b_2(Z,W, \tilde\alpha_2)\mid A=a_3, W]\}\notag \\ 
    & = \arg\min_{\tilde\alpha_2}\E\{\tilde\alpha_2(A, Z, W)^2 - 2\alpha_1 (A, W) b_2(Z,W, \tilde\alpha_2)\} \label{eq:alpha2},
\end{align}
where the last equality can be derived by defining $q_1(a, w) = \E[b_2(Z, W; \tilde \alpha_2) \mid A = a, W = w]$ and noticing
\begin{align*}
   & \E \{ \E[b_2(Z,W, \tilde\alpha_2)\mid A=a_3, W]\} = \E[q_1(a_3, W)] = E[b_1(W; q_1)] = \E[\alpha_1(A, W) q_1(A, W)] \\
   & = \E \{ \E[\alpha_1(A, W) b_2(M, W; \tilde \alpha_2) \mid A = a_3, W ] \} = \E [ \alpha_1(A, W) b_2(M, W; \tilde \alpha_2)].
\end{align*}
As for $\alpha_3$, we have
\begin{align}
    \alpha_3  & =  \arg\min_{\tilde\alpha_3}\E[\tilde\alpha_2(A, Z, M, W)^2 - 2\E\{\E [b_3(Z,M,W;\tilde \alpha_3)\mid A=a_2,Z, W ] \mid A=a_3, W \} ] \notag \\ 
    & = \arg\min_{\tilde\alpha_3}\E[\tilde\alpha_3(A, Z, M, W)^2 - 2\alpha_2 (A, Z, W) b_3(Z,M,W, \tilde\alpha_3)] \label{eq:alpha3},
\end{align}
while for the last equality, we define $q_2 (a, z, w) = \E \{b_3(Z,M,W;\tilde \alpha_3)\mid A = a, Z = z, W = w \}$ and we can obtain
\begin{align*}
    & \E\{\E[\E \{b_3(Z,M,W;\tilde \alpha_3)\mid A=a_2,Z, W \} \mid A=a_3, W] \} = E \{ E[ q_2(a_2, Z, W) \mid A = a_3, W] \} \\ 
    & = \E [\alpha_2(A, Z, W) q_2(A, Z, W) ] = \E [ \E \{ \alpha_2(A, Z, W) b_3(Z, M, W; \tilde \alpha_3) \mid A, Z, W\}] \\ 
    & =  \E \{ \alpha_2(A, Z, W) b_3(Z, M, W; \tilde \alpha_3) \}
\end{align*} $\square$

\end{document}